\documentclass[final,hidelinks,onefignum,onetabnum]{siamart251216}
\usepackage{amsmath}
\usepackage{amssymb}
\usepackage{lipsum}
\usepackage{amsfonts}
\usepackage{graphicx}
\usepackage{algorithmic}
\usepackage{booktabs}
\usepackage{multirow}
\PassOptionsToPackage{url}{hyperref}
\usepackage{enumitem}
\usepackage{bbm}
\usepackage{bm}
\usepackage{tikz, tikz-cd}
\usetikzlibrary{positioning, arrows.meta}
\usepackage[normalem]{ulem}
\usepackage[section]{placeins}  
\ifpdf
  \DeclareGraphicsExtensions{.eps,.pdf,.png,.jpg}
\else
  \DeclareGraphicsExtensions{.eps}
\fi

\makeatletter
\let\sdifp@paragraph\@paragraph
\makeatother

\newsiamremark{hypothesis}{Hypothesis}
\crefname{hypothesis}{Hypothesis}{Hypotheses}
\newsiamthm{claim}{Claim}
\crefname{claim}{Claim}{Claims}
\newsiamthm{exm}{Example}
\crefname{exm}{example}{examples}
\Crefname{exm}{Example}{Examples}
\crefname{figure}{Fig.}{Figs.}
\Crefname{figure}{Fig.}{Figs.}
\crefname{proposition}{Proposition}{Propositions}
\Crefname{proposition}{Proposition}{Propositions}
\crefname{corollary}{Corollary}{Corollaries}
\Crefname{corollary}{Corollary}{Corollaries}
\crefname{theorem}{Theorem}{Theorems}
\Crefname{theorem}{Theorem}{Theorems}
\newsiamremark{remark}{Remark}
\crefname{remark}{Remark}{Remarks}

\newtheorem{assumption}{Assumption}[section]

\newenvironment{Proof}
  {\par\noindent\textbf{Proof.}}
  {\qed\par}

\providecommand{\R}{\mathbb R}

\providecommand{\Q}{\mathcal Q}

\headers{Stochastic Dimension Implicit Functional Projections}{Z. Liang, H. Gao, M. Xiao}

\title{Stochastic Dimension Implicit Functional Projections for Global Integral Conservation in High-Dimensional PINNs
\thanks{Submitted to the editors \today.
\funding{This work was supported by the National Natural Science Foundation of China (NO. 12202157).}}}

\author{Zhangyong Liang\thanks{National Center for Applied Mathematics, Tianjin University, Tianjin, 300072, China (\email{zyliang1994@tju.edu.cn})}.
\and Huanhuan Gao\thanks{School of Mechanical and Aerospace Engineering, Jilin University, Changchun 130025, China (\email{gao\_huanhuan@jlu.edu.cn}).}
\and Manyu Xiao\thanks{Xi'an Key Laboratory of Scientific Computation and Applied Statistics, School of Mathematics and Statistics, Northwestern Polytechnical University, Youyixi Road 127, Xi'an 710072, China (\email{manyuxiao@nwpu.edu.cn}).}}

\usepackage{amsopn}

\makeatletter
\providecommand{\headerps@out}[1]{}
\makeatother

\hypersetup{
  pdftitle={Stochastic Dimension Implicit Functional Projections for Global Integral Conservation in High-Dimensional PINNs},
  pdfauthor={Zhangyong Liang, Huanhuan Gao, Manyu Xiao}}
\date{}

\begin{document}
\maketitle

\begin{abstract}
Enforcing global integral constraints in neural partial differential equation (PDE) solvers requires numerical quadrature, while high-dimensional differential operators can make reverse-mode automatic differentiation (AD) costly. We propose the Stochastic Dimension Implicit Functional Projection (SDIFP) framework for enforcing prescribed first and second spatial moments under a positive weighted quadrature rule. A global affine correction of the network output determines two scalar coefficients from the quadrature moments. For positive target variance and nondegenerate raw quadrature variance, the corrected nodal vector is the unique nearest point on the empirical two-moment constraint set in the weighted norm. The prescribed moments are exact for the calibration rule in exact arithmetic with an inactive variance floor. Continuum moment errors remain subject to quadrature error and may be amplified near degeneracy. We combine this correction with stochastic operator-subset sampling for decomposable linear differential operators. Independent residual and derivative samples, together with conditionally unbiased gradients of all coefficient-dependent terms, give an unbiased estimator of the specified quadrature-based squared-residual gradient. Reusing the same operator subset generally introduces bias. Detached quadrature separates forward moment evaluation from coefficient-gradient reconstruction. Once coefficients are available at a specified time, pointwise evaluation requires one network evaluation and an affine correction. Algebraic diagnostics and historical plotting examples illustrate moment matching, structured coefficient-gradient reconstruction, and operation-count models.
\end{abstract}



\section{Introduction}\label{sec:introduction}

Neural network-based methods are increasingly used to solve and model partial differential equations (PDEs) \cite{brunton2020machine, kovachki2023neural}. Physics-informed neural networks (PINNs) \cite{raissi2019physics} are a common example. They parameterize the PDE solution continuously, require little labeled data, and can be trained on mesh-free collocation points. Standard PINN losses enforce the PDE residual, initial data, and boundary data through penalty terms. This soft enforcement is flexible, but it may allow violations of invariants that are important for long-time simulation \cite{rathore2024challenges}.

In applications where global invariants are essential, it is useful to impose integral constraints directly on the model output. Examples include total mass and selected quadratic moments. Treating these quantities only as penalty terms can introduce tuning sensitivity and sampling-dependent errors. Hard or algebraic enforcement is therefore attractive when the target invariant is known and compatible with the boundary conditions.

Imposing global integral constraints in a mesh-free neural solver is still challenging. Methods such as PINN-Proj \cite{baez2025guaranteeing} project spatial outputs onto a constraint set using numerical quadrature. Quadrature nodes and residual collocation points serve different purposes: random residual sampling is compatible with a fixed calibration rule. Replacing the calibration rule by a fresh random batch changes the empirical constraint. The resulting coefficients may depend on sampling noise.

Differentiable optimization layers such as $\Pi$net \cite{grontas2026pinet} provide another route. They use operator splitting methods, for example Douglas-Rachford iterations \cite{dao2023douglas}, together with implicit differentiation \cite{dontchev2009implicit}. Their standard convergence theory is strongest for closed convex feasible sets. For a scalar field $u$ on a spatial domain $\mathcal X$ and a prescribed $C>0$, the quadratic moment equality $\int_{\mathcal{X}} u^2 \, dx = C$ defines a nonconvex feasible set requiring a tailored convergence analysis. Discrete projections also couple many spatial points during backpropagation \cite{grazzi2020iteration}, which can be costly in high-dimensional settings.

Scaling constrained neural PDE solvers to high dimensions also creates a bottleneck in high-order spatial differential operators. Reverse-mode automatic differentiation across many spatial dimensions can require substantial memory \cite{yao2020pyhessian}, particularly for high-order operators. Several efforts have been proposed to mitigate this differential curse of dimensionality. Stochastic Dimension Gradient Descent (SDGD) \cite{hu2024tackling} randomizes over input dimensions. At each iteration, derivatives are computed only for a sampled subset of dimensions. In \cite{hu2024hutchinson,lai2022regularizing,hu2024score}, the Hutchinson trace estimator (HTE) \cite{hutchinson1989stochastic} estimates traces of Hessians or Jacobians with respect to inputs. Others choose to bypass AD completely to reduce the complexity of computation. In \cite{pang2020efficient}, the finite difference method is used for estimating the Hessian trace. Randomized smoothing \cite{he2023learning,hu2023bias} uses an expectation over Gaussian perturbations as the ansatz. Its derivatives can then be expressed through Stein's identity \cite{stein1981estimation}. More recently, the Stochastic Taylor Derivative Estimator (STDE) \cite{shi2024stochastic} uses Taylor-mode automatic differentiation to estimate contractions of high-order derivative tensors. Global integral enforcement complements these approaches to reducing differentiation cost.

\begin{figure}[!htbp]
    \centering
    \includegraphics[width=\linewidth]{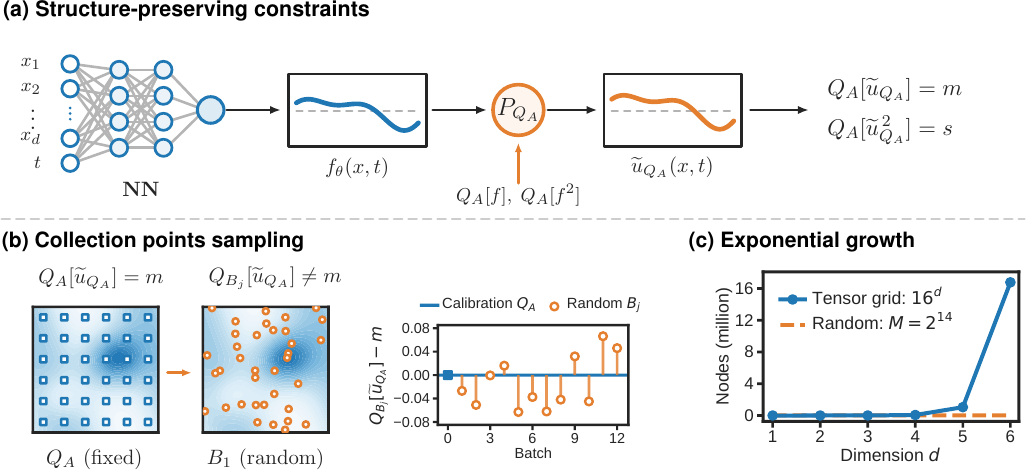}
    \caption{Motivation for SDIFP. (a) Moment calibration of a neural field. (b) Integral mismatch between a fixed calibration rule $Q_A$ and random evaluation batches $B_j$, with prescribed normalized mean $m$. (c) Exponential tensor-grid node growth compared with a fixed random-sampling budget.}
    \label{fig:sdifp_motivation}
\end{figure}

Quadrature-based hard conservation constraints are commonly calibrated using a fixed set of spatial nodes. Fig.~\ref{fig:sdifp_motivation} illustrates the resulting support and scaling issues. Panel (a) shows a neural field corrected to satisfy prescribed moments on its calibration rule. Panel (b) shows that evaluating the same corrected field on fresh random points introduces an empirical integral discrepancy. Thus a change in sampling support requires explicit control of the relation between calibration and evaluation. Panel (c) highlights the cost of a tensor-product calibration grid: with $n$ nodes per coordinate in $d$ spatial dimensions, the total is $n^d$. Evaluating global moments on that grid therefore entails exponentially growing node-evaluation work for fixed $n>1$, making this implementation of hard conservation constraints costly in high dimensions. This motivates a formulation that combines global moment calibration on weighted point sets with sampled differentiation.

\FloatBarrier

To address this issue, we propose the \textbf{S}tochastic \textbf{D}imension \textbf{I}mplicit \textbf{F}unctional \textbf{P}rojection (\textbf{SDIFP}) framework. Let $u_{\mathrm{raw}}(x;\theta)$ be the unconstrained network output at spatial point $x$, with network parameters $\theta$; time is suppressed here. SDIFP applies an affine correction
\[
    \tilde{u}(x)=\alpha_\Q\, u_{\mathrm{raw}}(x;\theta)+\beta_\Q.
\]
Here $\Q$ denotes the calibration quadrature rule, and $\alpha_\Q$ and $\beta_\Q$ are scalar coefficients that depend on time and network parameters (Section~\ref{sec:sdifp}). They match prescribed first and second moments under that rule. In SDIFP, ``implicit refers to the moment equations defining the coefficients, and ``functional refers to their affine extension to arbitrary spatial points. The center--scale--shift form is shared with PINN-Proj under uniform quadrature. Our focus is its weighted geometry and its combination with sampled coefficient gradients and operator terms.

The paper makes the following contributions:
\begin{itemize}[leftmargin=1.5em]
    \item We propose SDIFP, a mesh-free affine projection framework that enforces prescribed first and second spatial moments with respect to a weighted quadrature rule, avoiding tensor-product nodal projections for the moment constraints.

    \item We establish the mathematical structure of the correction: in the nondegenerate case, SDIFP is the nearest-point projection in the weighted quadrature norm onto the empirical two-moment shell. We further characterize feasibility, degeneracy, variance-floor regularization, and continuum quadrature errors.

    \item We develop a stochastic operator-sampling training strategy for decomposable high-dimensional linear differential operators. We state the precise conditions under which the estimator is unbiased for the quadrature-based residual-gradient target and identify the shared-subset mode as biased in general.

    \item We validate algebraic moment diagnostics and structured coefficient-gradient reconstruction, confirming high-precision gradient convergence ($10^{-6}$ to $10^{-8}$) alongside substantial reverse-graph reduction.
\end{itemize}

Section \ref{sec:preliminaries} reviews the theoretical bottlenecks of explicit discrete projections and high-dimensional operators. Section \ref{sec:method} develops the SDIFP framework, covering the projection derivation and the doubly stochastic gradient estimator. Section \ref{sec:theory} establishes projection optimality, support-transfer identities, stability, and conditional gradient unbiasedness. Section \ref{sec:experiments} reports quadrature-level integral errors and computational diagnostics. Section \ref{sec:conclusion} collects the conclusions.

\section{Related Work}\label{sec:work}

\textbf{Soft Constraints on PINNs.} Soft penalty terms for constraint violations in the objective function are one approach to incorporate constraints in neural networks \cite{marquez2017imposing}, and have been adopted to solve parametric constrained optimization problems \cite{tuor2021neuromancer}, and partial differential equations through physics-informed neural networks \cite{erichson2019physics, raissi2019physics}. A soft constraint can be used to enforce the conservation law in a PINN by adding a loss term that penalizes the difference between the conserved quantity in the current prediction and the ground truth \cite{son2022alpinn}. Training can either be done entirely with this modified loss function \cite{fang2022data}, or in two stages where the modified loss function is used in a second stage of training \cite{lin2022two}. Creating soft constraints in this way can also be used to enforce conservation of flux between two neighboring discrete subdomains \cite{jagtap2020conservative}. Their constraint accuracy depends on optimization and inference conditions. Penalty weights affect constraint accuracy \cite{ping2023adaptive}. This motivates several alternatives. First, the structure of the constraint set can be exploited to design more efficient algorithms. Second, treating constraints softly may significantly alter the problem solution regardless of tuning \cite{grontas2024osp}. Third, softened formulations can be ill-posed even for linear programs \cite{basir2022pecann}.

\textbf{Hard Constraints on PINNs.} 
Hard constraints have been used in PINNs to strictly enforce initial and boundary conditions \cite{xiao2024hard}. 
KKT-hPINN uses KKT conditions to construct an untrainable projection for hard linear constraints \cite{chen2024physics}.
To circumvent the shortcomings of soft constraints, hard-constrained neural networks aim to enforce constraints on the NN output by design. 
\cite{frerix2020homogeneous} address linear homogeneous inequality constraints by parameterizing the feasible set. 
RAYEN \cite{tordesillas2023rayen} enforces convex constraints by scaling the line segment between infeasible points and a fixed point in the feasible set's interior. 
These constructions use offline preprocessing to support rapid inference on fixed constraint sets.
\cite{min2024hard} propose a closed-form expression for polyhedral constraints and employ CVXPYlayers \cite{agrawal2019differentiable} for general convex sets. 
\cite{cristian2023end} orthogonally project the NN output or intermediate layers by unrolling Dykstra's algorithm \cite{boyle1986method}.
DC3 \cite{donti2021dc3} introduces an equality completion and inequality correction procedure. \cite{lastrucci2025enforce} impose non-linear equality constraints by recursively linearizing them. Lagrangian and augmented Lagrangian approaches \cite{fioretto2021lagrangian, park2023self} are considered for general non-convex constraints. \cite{kratsios2021universal} consider a probabilistic sampling approach. LinSATNet \cite{wang2023linsatnet} imposes non-negative linear constraints, relaxed by GLinSAT \cite{zeng2024glinsat}. A projection-based method applies multiple Newton iterations after each training step to guarantee conservation \cite{cardoso2025exactly}. AT-PINN-HC \cite{cheung2025atpinn} introduces hard-constraint strategies with various auxiliary functions for structural vibration problems. HCP-PINN \cite{wang2026hcpinn} extends hard constraint methods to strongly nonlinear PDEs.

\textbf{Learning Conservation Laws.} Conservation laws can be mapped to hard constraints by incorporating them directly into the structure of neural networks. Hamiltonian \cite{greydanus2019hamiltonian} and Lagrangian \cite{cranmer2019lagrangian} formulations provide structural representations of dynamical systems. Hard constraints can also be created by parameterizing a divergence-free vector field \cite{richter2022neural}. Neural operators have been applied to enforce conservation using the continuity equation \cite{liu2023harnessing}.

\section{Preliminaries}\label{sec:preliminaries}

In this section, we review the fundamental concepts of PINNs, the mathematical formulation of global conservation laws, and recent explicit projection methods designed to enforce these constraints.

\subsection{Physics-informed neural networks and soft constraints}\label{sec:pinn_sc}

We consider a physical system governed by a general PDE over a spatial domain $\mathcal{X} \subset \mathbb{R}^d$ and a time domain $\mathcal{T}$:
\begin{equation}
    \frac{\partial u}{\partial t} + \mathcal{N}[u] = 0, \quad x \in \mathcal{X}, \; t \in \mathcal{T},
\end{equation}
where $u(x, t)$ is the continuous state variable and $\mathcal{N}[\cdot]$ is a differentiable nonlinear spatial operator. A standard PINN approximates the latent solution via a neural network $u_\theta(x, t)$ parameterized by weights $\theta$. The network is trained by minimizing a composite loss function $\mathcal{L}$, which penalizes both data discrepancies and PDE residuals.

To enforce global conservation laws, a conventional approach, referred to as PINN with soft constraints (PINN-SC), incorporates an additional penalty term into the objective function. The resulting integral accuracy depends on the penalty weights and optimization error.

\subsection{Global conservation laws and integral invariants}\label{sec:conservation_laws}

Many fundamental PDEs naturally describe the conservation of physical quantities. A canonical conservative form is expressed as
\begin{equation}
    \frac{\partial u}{\partial t} + \nabla \cdot \mathbf{F}(u) = 0.
\end{equation}
Integrating the conservation law over the spatial domain $\mathcal{X}$ yields the total quantity over the domain. For non-conserved systems, for example systems with sources, sinks, or open boundaries, the target quantity $c(t)$ may evolve in time and must be specified separately.

In this work, we focus on two global integral diagnostics over the spatial domain $\mathcal{X}$. The linear integral and quadratic $L^2$-type integral are
\begin{equation}
    c_1(t) = \int_{\mathcal{X}} u(x, t) \, dx, \quad c_2(t) = \int_{\mathcal{X}} u^2(x, t) \, dx.
\end{equation}
For a given PDE, these quantities are used as constraints only when the PDE and boundary conditions make them meaningful invariants or when a prescribed reference trajectory is available.

\subsection{Explicit discrete projections}\label{sec:explicit_projections}

Because analytic integration of neural-network outputs is generally intractable, recent explicit projection methods such as PINN-Proj \cite{baez2025guaranteeing} evaluate the integrals by numerical quadrature. Given a uniform grid with spatial discretization size $\Delta x$ and $N$ total grid points, the integrals are approximated by deterministic Riemann sums. PINN-Proj forces the discrete network outputs to respect the integral constraints by solving a Euclidean constrained optimization problem. For feasible, nondegenerate two-moment constraints, this gives an analytical projection of the nodal vector.

Extending the framework to arbitrary spatial dimension $d$, we consider $u_\theta(\mathbf{x}, t)$ over $\mathcal{X} \subset \mathbb{R}^d$, with $\mathbf{x} = (x^{(1)}, x^{(2)}, \dots, x^{(d)})$. The linear integral in $d$ dimensions takes the form
\begin{equation}
    \hat{c}_1(t) = \int_{\mathcal{X}} u_\theta(\mathbf{x}, t) \, d\mathbf{x} \approx \sum_{i=1}^{N} u_\theta(\mathbf{x}_i, t) \, \Delta V,
\end{equation}
where $\Delta V = (\Delta x)^d$ is the $d$-dimensional volume element of the uniform grid.

\section{Methodology}\label{sec:method}

This section develops the SDIFP correction for two global scalar moments. We describe the weighted nodal projection, its affine extension, and stochastic operator sampling for decomposable linear differential operators. Boundary compatibility is addressed after the estimator construction.

\subsection{Motivation}\label{sec:motivation}

Recent methods impose integral constraints by projecting network predictions onto a feasible set. When this projection is formulated on discrete spatial evaluations, the method inherits the structure and cost of the quadrature rule.

\subsubsection{Explicit discrete projections and grid dependency}

Methods such as PINN-Proj formulate conservation constraints as discrete Euclidean projections. Evaluating and differentiating a projection on a large calibration set couples the quadrature-node evaluations. If this operation remains inside the automatic differentiation graph, memory use grows with the quadrature size.

\begin{figure}[!htbp]
    \centering
    \includegraphics[width=\linewidth]{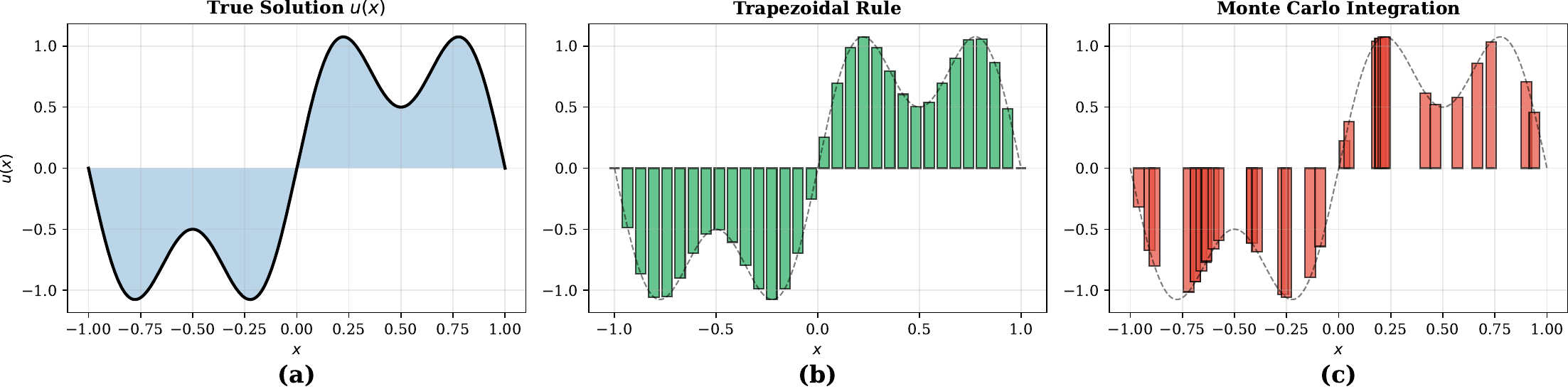}
    \vspace{-1.5em}
    \caption{Illustration of spatial integration: (a) a continuous field, (b) trapezoidal quadrature on a fixed grid, and (c) Monte Carlo sampling.}
    \label{fig:integral_compare}
\end{figure}

Fig.~\ref{fig:integral_compare} illustrates fixed-grid and Monte Carlo spatial quadrature. Calibration on one rule enforces that rule's empirical moments. Evaluation on another rule introduces a quadrature discrepancy for both PINN-Proj and SDIFP.

\subsubsection{Implicit optimization layers and nonconvex moment constraints}

Architectures such as $\Pi$net introduce a differentiable optimization layer for projecting raw network outputs onto a feasible set. Standard Douglas-Rachford convergence guarantees apply to suitable convex settings. Quadratic moment equalities with positive target require a separate nonconvex analysis. Differentiating a dense spatial projection couples the calibration-node evaluations.

\subsection{Stochastic dimension implicit functional projection (SDIFP)}\label{sec:sdifp}

We use the affine extension of the weighted nodal projection to enforce global first- and second-moment constraints.

\begin{figure}[!htbp]
    \centering
    \includegraphics[width=\linewidth]{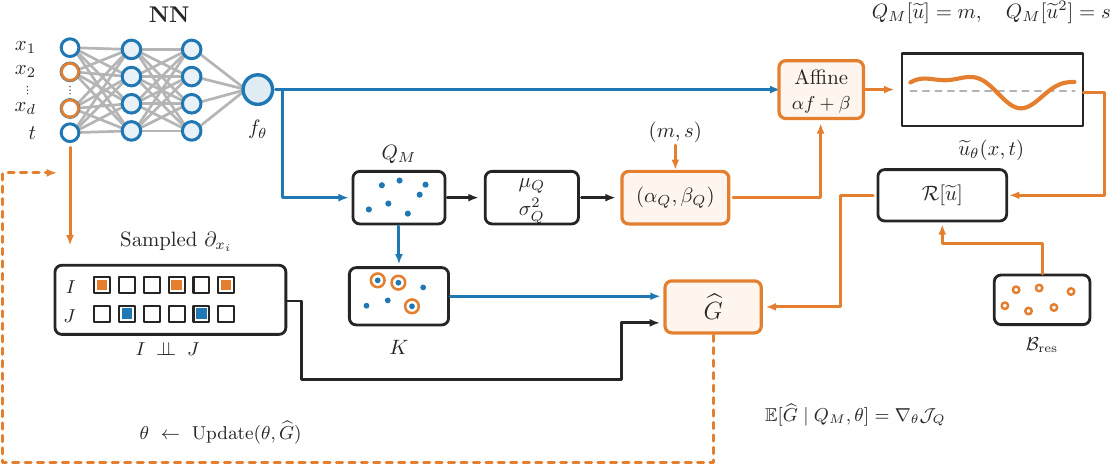}
    \caption{SDIFP framework. Full-rule moment calibration determines the affine correction, and independent node and operator sampling provide the stochastic parameter update.}
    \label{fig:sdifp_framework}
\end{figure}

Fig.~\ref{fig:sdifp_framework} summarizes the SDIFP computation. The raw network output is calibrated by a global affine correction whose coefficients are computed on the full quadrature rule $Q_M$. A node subset $K$ reconstructs the coefficient gradients, while independent operator subsets $I$ and $J$ supply the derivative and residual factors of the stochastic update. Residual collocation points enter through the corrected field. The definitions and independence conditions for these sampling steps are developed below.

We define the SDIFP map as a nearest-point projection onto an empirical first- and second-moment shell in a finite-dimensional weighted space. The affine formula then extends the corrected nodal values to the spatial domain.

\subsubsection{Setting and weighted quadrature norm}

Let $\mathcal{X}\subset\R^d$ have finite positive measure, and let $d$ denote the spatial dimension. Write
\[
    \mathcal{I}[g]=\frac{1}{|\mathcal{X}|}\int_\mathcal{X} g(x)\,dx,
\]
for the normalized continuum integral. At a fixed time $t$, let $f(x)=u_{\rm raw}(x,t;\theta)$ be the raw network output. Let $\{(x_m,w_m)\}_{m=1}^M$ be a quadrature rule with $M\ge2$, $w_m>0$ and $\sum_{m=1}^M w_m=1$. Define
\[
    {\mathcal Q}_M[g]=\sum_{m=1}^M w_m g(x_m),\qquad
    \langle g,h\rangle_\Q={\mathcal Q}_M[gh],\qquad
    \|g\|_\Q^2={\mathcal Q}_M[g^2].
\]

The weighted nodal norm is defined on $\boldsymbol g=(g(x_1),\ldots,g(x_M))\in\mathbb R^M$, with $W=\operatorname{diag}(w_1,\ldots,w_M)$ and $\|\boldsymbol g\|_\Q^2=\boldsymbol g^T W\boldsymbol g$. Its extension to arbitrary functions is a seminorm. Below, $g$ also denotes its nodal vector.

The prescribed normalized targets, independent of $\theta$, are
\[
    m(t)=\bar c_1(t)=\frac{c_1(t)}{|\mathcal{X}|},\qquad
    s(t)=\bar c_2(t)=\frac{c_2(t)}{|\mathcal{X}|},\qquad
    V_c(t)=s(t)-m(t)^2.
\]

We suppress $t$ and $\theta$ in moment notation when they are fixed. For the raw field define the weighted empirical mean, second moment, and variance
\[
    \mu_\Q={\mathcal Q}_M[f],\qquad
    \mu_{2,\Q}={\mathcal Q}_M[f^2],\qquad
    \sigma_\Q^2={\mathcal Q}_M[(f-\mu_\Q)^2]={\mathcal Q}_M[f^2]-\mu_\Q^2.
\]

We write $\sigma_\Q=\sqrt{\sigma_\Q^2}$ for the empirical standard deviation. The empirical moment feasible set in the nodal space is
\begin{equation}\label{eq:shell}
    \mathcal{C}_\Q(t)=\{v\in\mathbb R^M:{\mathcal Q}_M[v]=m(t),\ {\mathcal Q}_M[v^2]=s(t)\}.
\end{equation}

\begin{assumption}[Target feasibility and raw nondegeneracy]\label{ass:feas}
The target moments satisfy $V_c(t)\ge 0$. If $V_c(t)>0$, assume $\sigma_\Q^2>0$.
\end{assumption}

\subsubsection{Continuous implicit functional projections}

\begin{figure}[!htbp]
    \centering
    \includegraphics[width=\linewidth]{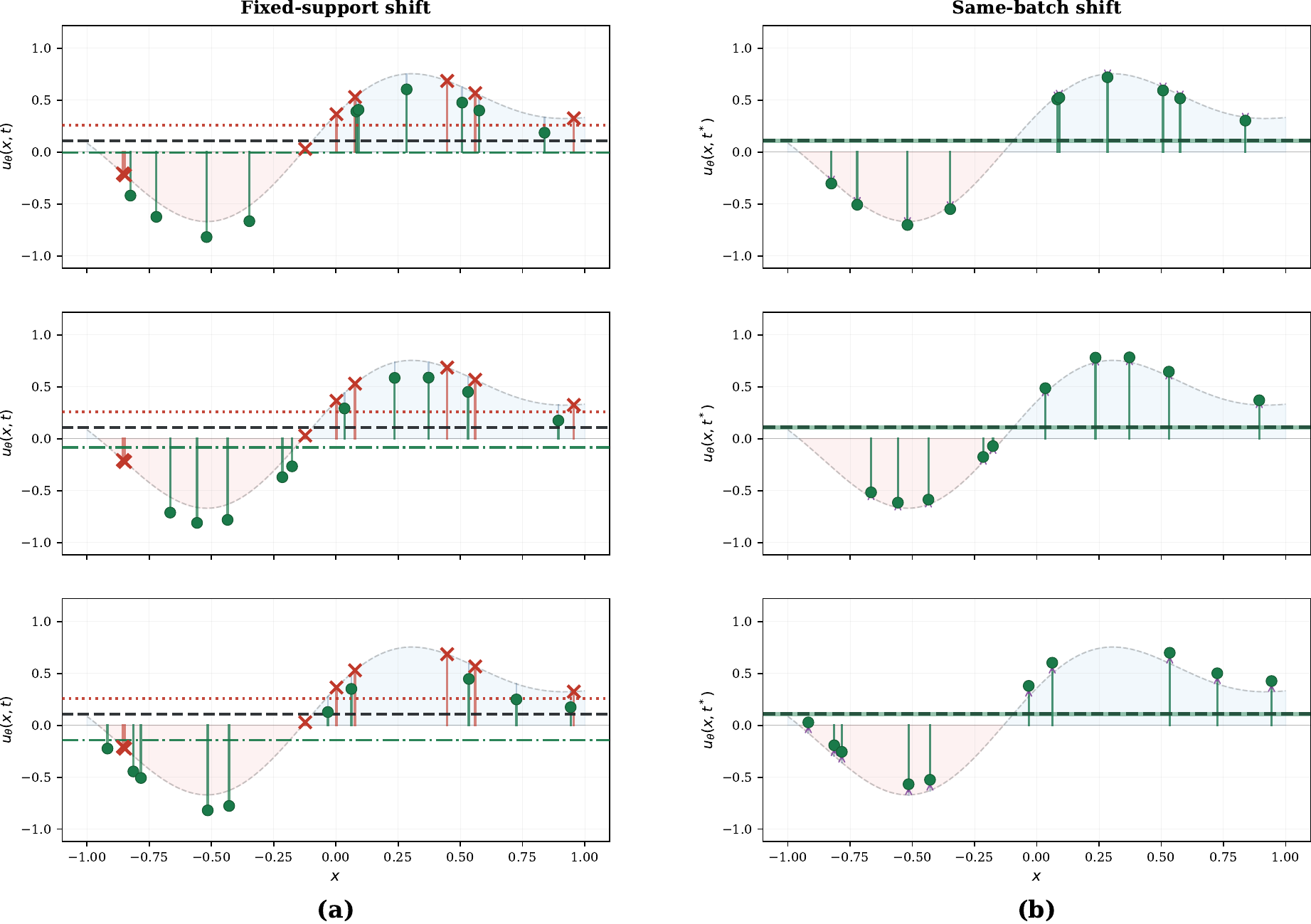}
    \vspace{-1.5em}
    \caption{First-moment shifts with fixed-support and same-batch calibration. Panel (a) applies a fixed-support correction to another batch. Panel (b) calibrates and evaluates on the same batch. Exact mean matching holds on each correction's calibration support.}
    \label{fig:sdifp_projection}
\end{figure}

Fig.~\ref{fig:sdifp_projection} illustrates the first-moment identity for fixed-support and same-batch calibration. The detached large-$M$ formulation uses its own calibration support.

In Fig.~\ref{fig:sdifp_projection}, the illustrative shift uses the same batch for calibration and evaluation. The full formulation instead computes $\alpha_\Q$ and $\beta_\Q$ from a detached large-$M$ quadrature set using the current $\theta$. For an equal-weight illustrative batch of size $n_b$, the same-batch identity is
\[
\frac{1}{n_b}\sum_{i=1}^{n_b}\tilde u(x_i,t)=\bar c_1(t).
\]
The full method enforces the corresponding identity on its calibration rule $\Q_M$.

For a finite support set $S$, write $\bar u(S)=|S|^{-1}\sum_{z\in S}u_{\mathrm{raw}}(z,t;\theta)$ for its sample mean.

\begin{table}[htbp]
\centering
\caption{Fixed-support and same-batch first-moment shifts, classified by calibration support.}
\label{tab:sdifp_vs_traditional_proj}
\small
\begin{tabular*}{\linewidth}{@{\extracolsep{\fill}}lll@{}}
\toprule & Fixed-support shift & Same-batch shift \\
\midrule
Integral support & Fixed $\{y_j\}$ & Current $\{x_i\}$ \\
Projection sites & Mini-batch $\{x_i\}$ & Same $\{x_i\}$ \\
Shift $\delta$ & $\bar{c}_1(t)-\bar{u}(\{y_j\})$ & $\bar{c}_1(t)-\bar{u}(\{x_i\})$ \\
Batch residual $e_{\mathrm{batch}}$ & $\bar{u}(\{x_i\})-\bar{u}(\{y_j\})$ & $0$ for the same batch \\
\bottomrule
\end{tabular*}
\end{table}

For either support choice, exact mean matching holds on the rule used to compute the shift. A fresh evaluation batch contributes its own sampling discrepancy.

\subsubsection{Affine quadrature projection onto the moment shell}

The affine correction is defined as
\begin{equation}
    \label{eq:affine_transform}
    \tilde{u}(x, t;\theta) = \alpha_\Q(t) \cdot u_{\rm raw}(x, t; \theta) + \beta_\Q(t).
\end{equation}
Here $u_{\rm raw}$ is the unconstrained network output and $\alpha_\Q(t),\beta_\Q(t)\in\mathbb R$ are the SDIFP coefficients given by \eqref{eq:exact_alpha_beta}.

The affine correction is the unique nearest nodal vector on the weighted moment shell under the nondegeneracy conditions of Theorem~\ref{th:projection}. Its proof and quantitative properties are given in Section~\ref{sec:theory}.

\begin{remark}[Projection space and affine extension]
The nearest-point property holds in the finite-dimensional weighted nodal space. Equation~\eqref{eq:affine_transform} extends the projected values through a global affine moment correction on $\mathcal X$.
\end{remark}

\begin{remark}[Variance floor]
For $\varepsilon>0$, the replacement $\sigma_\Q^2\mapsto\max(\sigma_\Q^2,\varepsilon)$ gives a positive denominator. The first moment remains exact in exact arithmetic. The second-moment defect is $\Q_M[\tilde u^2]-s=V_c(\sigma_\Q^2/\max(\sigma_\Q^2,\varepsilon)-1)$. Floating-point errors affect both moments.
\end{remark}

\begin{remark}[Continuum moment error]
With an inactive variance floor, the continuum moment errors equal the quadrature errors of the corrected field and its square. For a fixed raw field $f\in L^4(\mathcal X)$ with positive continuum variance and i.i.d. uniform nodes of equal weight, both errors are $O_p(M^{-1/2})$ (Appendix~\ref{app:continuum}). This rate describes convergence in probability for a fixed raw field. Assessing a field trained on the calibration sample requires independent evaluation quadrature. Deterministic quasi-Monte Carlo rates depend on discrepancy and integrand-variation assumptions.
\end{remark}

\begin{remark}[Degenerate cases]
If $V_c=0$, every feasible vector satisfies $v(x_m)=m$ on all positive-weight nodes. For $V_c>0$ and $\sigma_\Q^2=0$, the raw nodal vector is constant and the affine scaling formula is undefined. Every feasible nodal vector then has the same distance from the raw vector, so selecting a minimizer requires a tie-breaking rule.
\end{remark}

The closed-form coefficients from Theorem~\ref{th:projection} are
\begin{equation}
\label{eq:exact_alpha_beta}
\alpha_\Q(t) = \sqrt{\frac{V_c(t)}{\sigma_\Q^2(t;\theta)}},
\qquad
\beta_\Q(t) = m(t) - \alpha_\Q(t) \mu_\Q(t;\theta).
\end{equation}
These are evaluated from the detached quadrature set $\{(x_m,w_m)\}_{m=1}^M$. Here a quadrature evaluation is \emph{detached} if it is excluded from the automatic differentiation graph; in practice this is achieved by applying a stop-gradient (or \texttt{detach}) operator. This removes the large quadrature set from reverse-mode AD.

\subsubsection{Doubly stochastic gradient estimator (DS-UGE)}

During network optimization, computing the gradient of the PDE residual loss requires differentiating through the correction. Applying the chain rule to \eqref{eq:affine_transform} yields:
\begin{equation}
    \label{eq:chain_rule}
    \nabla_\theta \tilde{u}(x, t) = \alpha_\Q(t) \nabla_\theta u_{\rm raw}(x, t; \theta) + u_{\rm raw}(x, t; \theta) \nabla_\theta \alpha_\Q(t) + \nabla_\theta \beta_\Q(t),
\end{equation}
The coefficients depend on $\theta$ through the empirical moments. Equation~\eqref{eq:jacobians} gives the partial derivatives with respect to those moments; their chain-rule composition gives the total parameter gradients below.
The partial derivatives of the coefficients with respect to the empirical moments are
\begin{equation}
    \label{eq:jacobians}
    \begin{aligned}
        \frac{\partial \alpha_\Q}{\partial \mu_\Q} &= \frac{\alpha_\Q \mu_\Q}{\sigma_\Q^2}, \quad
        &\frac{\partial \alpha_\Q}{\partial \mu_{2,\Q}} &= -\frac{\alpha_\Q}{2 \sigma_\Q^2}, \\
        \frac{\partial \beta_\Q}{\partial \mu_\Q} &= -\alpha_\Q - \mu_\Q \frac{\partial \alpha_\Q}{\partial \mu_\Q}, \quad
        &\frac{\partial \beta_\Q}{\partial \mu_{2,\Q}} &= - \mu_\Q \frac{\partial \alpha_\Q}{\partial \mu_{2,\Q}},
    \end{aligned}
\end{equation}
Using the empirical moments defined above, the total gradients are
\[
\nabla_\theta\alpha_\Q
=\frac{\alpha_\Q\mu_\Q}{\sigma_\Q^2}\nabla_\theta\mu_\Q
-\frac{\alpha_\Q}{2\sigma_\Q^2}\nabla_\theta\mu_{2,\Q},\qquad
\nabla_\theta\beta_\Q=-\mu_\Q\nabla_\theta\alpha_\Q-\alpha_\Q\nabla_\theta\mu_\Q.
\]
These formulas assume an inactive variance floor. On the branch $\sigma_\Q^2<\varepsilon$, $\nabla_\theta\alpha_{\Q,\varepsilon}=0$ and $\nabla_\theta\beta_{\Q,\varepsilon}=-\alpha_{\Q,\varepsilon}\nabla_\theta\mu_\Q$. At equality, the maximum is nonsmooth and a branch convention must be specified. Because the full quadrature pass is detached, coefficient gradients must be reconstructed. Appendix~\ref{app:unbiased_subsampling} gives the weighted estimators for a uniformly sampled index set $K\subset\{1,\ldots,M\}$.

Conditional on the calibration rule and current parameters, coefficient-gradient subsampling preserves the forward coefficients and reduces the number of quadrature nodes retained for parameter differentiation. Resampling the calibration rule changes the forward coefficients through the new empirical moments.

At a fixed time and fixed trained parameters, the coefficients may be precomputed. Each spatial query then requires one network evaluation and an affine multiply-add. The coefficients remain functions of time, and a new time requires evaluation or a specified approximation of them.

The affine correction remains time dependent under parameter-gradient detachment. For fixed spatial quadrature nodes and weights,
\[
\partial_t\tilde u=\alpha_\Q f_t+\alpha_\Q'f+\beta_\Q',\qquad
\partial_{tt}\tilde u=\alpha_\Q f_{tt}+2\alpha_\Q'f_t+\alpha_\Q''f+\beta_\Q''.
\]
Residual evaluation includes the coefficient time derivatives and their parameter gradients. Appendix~\ref{app:time_derivative} gives the moment differentiation formulas. Omitting these terms defines a different residual.

\subsubsection{Stochastic operator sampling}

High-dimensional PDE operators create an additional AD memory cost. We consider operators that can be partitioned into a high-dimensional linear principal part $\mathcal{A}_{\mathrm{lin}}$ and a nonlinear or lower-dimensional component $\mathcal{N}_{\mathrm{nonlin}}$. We write $\mathcal{A}[u] = \mathcal{A}_{\mathrm{lin}}[u] + \mathcal{N}_{\mathrm{nonlin}}[u]$.

Write the decomposable linear spatial part as $\mathcal A_{\mathrm{lin}}=\sum_{k=1}^{N_{\mathcal L}}A_k$, where $N_{\mathcal L}$ is the number of operator terms. Fully unrolling the reverse-mode graph for all $\nabla_\theta A_k[\tilde{u}_\theta]$ can incur an $O(N_{\mathcal{L}})$ memory cost for high-order terms. We therefore sample operator subsets only for decomposable linear components.

The subset $J\subset\{1,\ldots,N_{\mathcal L}\}$ estimates the linear residual, and $I$ indexes the linear terms used for its parameter-gradient estimate. Let $\mathcal B_{\mathrm{res}}$ be the residual batch of space-time points. Subset sizes satisfy $1\leq |I|,|J|\leq N_{\mathcal L}$.

\begin{assumption}[Independent linear operator sampling]\label{ass:op}
Conditional on the calibration rule and current parameters, $I$, $J$, $K$, and the residual points are mutually independent. The operator subsets $I$ and $J$ are uniform subsets of prescribed sizes, each sampled without replacement. All required second moments exist, and differentiation may be interchanged with the residual expectation.
\end{assumption}

For a linear spatial operator, the affine map satisfies $A_k[\tilde{u}_\theta]=\alpha_\Q A_k[u_{\mathrm{raw}}]+\beta_\Q A_k[\mathbf 1]$, where $\mathbf 1(x)\equiv 1$. With residual mini-batch $\mathcal{B}_{\mathrm{res}}$, the stochastic update involves the sampled operator terms. Let
\[
    r(x,t;\theta)=\sum_{k=1}^{N_{\mathcal L}}A_k[\tilde{u}](x,t)+\mathcal{N}_{\mathrm{nonlin}}[\tilde{u}](x,t)-R(x,t),
\]
where $\mathcal N_{\mathrm{nonlin}}$ collects the unsampled remainder, including temporal derivatives where required, and $R$ is the source. Its forward value is assumed exact for this estimator. Consider the space-time objective
\[
    \mathcal{J}_{\mathrm{PDE}}(\theta)=\frac12\mathbb E_{t\sim\rho_t}\mathbb E_{x\sim\rho_x}\left[r(x,t;\theta)^2\right],
\]
where $\rho_t$ and $\rho_x$ are the time and spatial sampling distributions. Initial and boundary penalties, when used, are added to this PDE loss and differentiated separately. The formulas below suppress $t$.
Under Assumption~\ref{ass:op}, define
\[
    \widehat r_J(x)=\frac{N_{\mathcal L}}{|J|}\sum_{j\in J}A_j[\tilde{u}](x)+\mathcal{N}_{\mathrm{nonlin}}[\tilde{u}](x)-R(x),
\]
\[
    \widehat d_{I,K}(x)=\frac{N_{\mathcal L}}{|I|}\sum_{i\in I}\widehat\nabla_\theta A_i[\tilde{u}](x)+\widehat\nabla_\theta \mathcal{N}_{\mathrm{nonlin}}[\tilde{u}](x),
\]
where $\widehat\nabla_\theta A_i[\tilde{u}]$ denotes the stochastic coefficient-gradient estimator of $\nabla_\theta A_i[\tilde{u}]$: by the chain rule \eqref{eq:chain_rule}, $\nabla_\theta A_i[\tilde{u}]$ contains the terms $\nabla_\theta\alpha_\Q$ and $\nabla_\theta\beta_\Q$, which are estimated via the unbiased estimators of Appendix~\ref{app:unbiased_subsampling} using $K$ sampled independently of $I$ and $J$. The hatted remainder gradient must likewise include all coefficient dependencies and be conditionally unbiased. The remainder gradient includes the same coefficient dependencies. For example, differentiating $\tilde u\,\partial_x\tilde u$ requires the gradients of both affine coefficients.

The composite doubly-stochastic gradient estimator is then
\begin{equation}
\label{eq:ds_uge_composite}
\widehat G(\theta) := \frac1{|\mathcal{B}_{\mathrm{res}}|}
\sum_{(x,t) \in \mathcal{B}_{\mathrm{res}}}
\widehat r_J(x)\,\widehat d_{I,K}(x).
\end{equation}

This partition reduces the number of retained linear operator terms. Gradient unbiasedness additionally requires conditional unbiasedness for every coefficient-dependent residual term.

The sampled linear-operator graph retains $|\mathcal B_{\mathrm{res}}||I|$ term evaluations; the full operator graph retains $|\mathcal B_{\mathrm{res}}|N_{\mathcal L}$. Coefficient-gradient reconstruction and unsampled remainder derivatives add separate costs. Total peak memory depends on graph reuse, network architecture, and operator order.

Reusing one subset, $I=J$, reduces repeated operator evaluations and generally introduces the covariance bias in \cref{cor:operator_reuse_bias}. Independent sampling of $I$ and $J$ gives the unbiased variant under the stated conditions. Independent draws may coincide; forcing distinct or disjoint draws generally introduces dependence. Training cost depends on both per-step work and the number of steps to convergence.

The complete procedure is summarized in \cref{alg:sdifp}, with $1\leq k\leq M$ and $1\leq i,j\leq N_{\mathcal L}$. Required time derivatives follow the PDE order and the regularity conditions of Appendix~\ref{app:time_derivative}. For $V_c>0$, the active floor preserves the first moment with the second-moment defect stated above; for $V_c=0$, the constant-field branch satisfies both targets.

\begin{algorithm}[!htbp]
\caption{SDIFP training with independent operator sampling}
\label{alg:sdifp}
\small
\begin{algorithmic}[1]
\REQUIRE Network $f_\theta$, parameters $\theta_0$, positive normalized rule $\Q_M$, feasible targets $m,s$, floor $\varepsilon>0$, residual batch size $B$, subset sizes $k,i,j$, learning rates $\eta_n$, and iteration count $N_{\rm it}$.
\ENSURE Parameters $\theta_{N_{\rm it}}$ and corrected field $\widetilde u_{\theta_{N_{\rm it}}}$.
\FOR{$n=0,\ldots,N_{\rm it}-1$}
    \STATE Set $\theta\gets\theta_n$; draw residual and initial/boundary batches.
    \FOR{each distinct batch time $t$}
        \STATE Evaluate $\mu_\Q=\Q_M[f_\theta]$, $\mu_{2,\Q}=\Q_M[f_\theta^2]$, and required time derivatives on the full rule.
        \STATE Set $\sigma_\Q^2=\mu_{2,\Q}-\mu_\Q^2$, $V_c=s-m^2$, $D_\Q=\max\{\sigma_\Q^2,\varepsilon\}$.
        \STATE Set $\alpha_\Q=\sqrt{V_c/D_\Q}$ and $\beta_\Q=m-\alpha_\Q\mu_\Q$.
        \STATE Use the exact branch for $\sigma_\Q^2>\varepsilon$ and the floored branch for $\sigma_\Q^2<\varepsilon$; at equality, apply the specified branch convention.
        \STATE Compute required coefficient time derivatives; retain forward values and release full-rule parameter graphs.
        \STATE Draw a uniform $k$-element subset $K_t\subseteq\{1,\ldots,M\}$ independently of the residual batch; set $f_q=f_\theta(x_q,t)$.
        \STATE $\widehat\nabla_\theta\mu_\Q\gets(M/k)\sum_{q\in K_t}w_q\nabla_\theta f_q$.
        \STATE $\widehat\nabla_\theta\mu_{2,\Q}\gets(2M/k)\sum_{q\in K_t}w_q f_q\nabla_\theta f_q$.
        \STATE Reconstruct coefficient gradients and mixed time--parameter derivatives using full-rule values and the same variance-floor branch.
    \ENDFOR
    \STATE Form $\widetilde u=\alpha_\Q f_\theta+\beta_\Q$ and its complete time derivatives.
    \STATE Draw independent uniform operator subsets $I,J$ of sizes $i,j$, independently of all $K_t$ and the residual batch.
    \FOR{$(x,t)\in\mathcal B_{\rm res}$}
        \STATE Evaluate $\widehat r_J(x,t)$ using full-rule coefficient values.
        \STATE Evaluate $\widehat d_{I,K}(x,t)$ with conditionally unbiased gradients of all coefficient-dependent terms.
    \ENDFOR
    \STATE $\widehat G_{\rm PDE}\gets B^{-1}\sum_{(x,t)\in\mathcal B_{\rm res}}\widehat r_J(x,t)\widehat d_{I,K}(x,t)$.
    \STATE Compute $\widehat G_{\rm IC/BC}$ from corrected initial/boundary losses with consistent coefficient derivatives.
    \STATE $\theta_{n+1}\gets\operatorname{Update}(\theta_n,\widehat G_{\rm PDE}+\widehat G_{\rm IC/BC};\eta_n)$.
\ENDFOR
\RETURN $\theta_{N_{\rm it}}$ and $\widetilde u_{\theta_{N_{\rm it}}}$, with full-rule coefficients evaluated at each query time.
\end{algorithmic}
\end{algorithm}

\textbf{Cost model.} The following costs are distinct:
\begin{itemize}[leftmargin=1.5em]
    \item Quadrature forward cost: $M$ network evaluations per coefficient update and per distinct time node. A batch with $n_t$ distinct times requires $n_t M$ evaluations without reuse. Required time derivatives add cost.
    \item Coefficient-gradient cost: $|K|$ quadrature-node derivative evaluations if subsampling is used, or $M$ for the exact quadrature gradient.
    \item Residual cost: $|\mathcal{B}_{\mathrm{res}}|$ residual points and sampled operator terms $|I|$, $|J|$.
    \item Reverse-mode AD memory: the sampled operator graph, the $|K|$-node coefficient-gradient graph at each time, and the unsampled remainder graph. Their sizes depend on network depth, width, input dimension, and operator order.
    \item Inference cost: one network evaluation plus one scalar multiplication and one scalar addition only when $\alpha_\Q$ and $\beta_\Q$ have been precomputed.
\end{itemize}

\begin{remark}[Compatibility with boundary conditions]
Let $\mathbf n$ denote the outward unit normal to $\partial\mathcal X$. The affine mapping preserves periodicity and homogeneous Neumann data satisfied by the raw network. For suitable flux laws, such as pure diffusion, homogeneous Neumann data also give physical zero flux. On a homogeneous Dirichlet boundary, the corrected value is $\beta_\Q(t)$, so preserving zero boundary values requires $\beta_\Q(t)=0$. A boundary-vanishing mask, such as $\tilde u=\alpha_\Q u_{\mathrm{raw}}+\beta_\Q\phi$ with $\phi|_{\partial\mathcal X}=0$, leads to a moment system containing mixed moments with $\phi$.
\end{remark}

The initial field is preserved only if its calibration moments match the prescribed initial targets. Even when $f(x,0)=u_0(x)$, using continuum targets with $\Q_M[u_0]\neq\mathcal I[u_0]$ or $\Q_M[u_0^2]\neq\mathcal I[u_0^2]$ can alter the initial data. Quadrature-consistent targets or an explicit initial-error assessment are therefore needed.

\begin{remark}[Numerical stability]
To prevent division by zero when the raw network is nearly flat, we replace $\sigma_\Q^2(t;\theta)$ by $\max(\sigma_\Q^2(t;\theta),\varepsilon)$ with $\varepsilon=10^{-8}$. This bounds the scaling factor by $\sqrt{V_c/\varepsilon}$. For $V_c>0$, an active floor preserves the first moment but introduces the second-moment defect stated above. If $V_c=0$, both target moments are still satisfied. The forward and backward branches must agree.
\end{remark}

\section{Theoretical analysis}\label{sec:theory}

We analyze the calibrated field, its change under a different integration
rule, and the stochastic gradient used to train it. Unless stated otherwise,
$Q:=\Q_M$ is a positive normalized quadrature rule held fixed in time and
independent of $\theta$. For reference, $\mu_\Q=Q[f]$ and $\sigma_\Q^2=Q[(f-\mu_\Q)^2]$.
Time is suppressed in the projection results.
The exact projection statements assume $V_c=s-m^2>0$ and
$\sigma_\Q^2>0$, with no active variance floor. The norm $\|\cdot\|_\Q$
acts on nodal vectors. Its affine extension defines the field away from the nodes.

\subsection{Weighted projection and calibration-support consistency}
\label{sec:theory_projection}

\begin{theorem}[Exact weighted nodal projection]\label{th:projection}\label{prop:exact_moments}\label{thm:quadrature_projection}
Let $M\geq2$, $w_q>0$, and $\sum_{q=1}^M w_q=1$. For any raw nodal
vector $f$ with positive variance, the correction
\begin{equation}\label{eq:projection}
 P_\Q f=m+\alpha_\Q(f-\mu_\Q),\qquad
 \alpha_\Q=\frac{\sqrt{V_c}}{\sigma_\Q},
\end{equation}
is the unique minimizer of $\tfrac12\|v-f\|_\Q^2$ over
$\mathcal C_\Q=\{v\in\mathbb R^M:Q[v]=m,\ Q[v^2]=s\}$.
In particular,
\begin{equation}\label{eq:theory_projection_exact}
 \begin{gathered}Q[P_\Q f]=m,\qquad Q[(P_\Q f)^2]=s,\\
 \|P_\Q f-f\|_\Q^2=(m-\mu_\Q)^2+(\sqrt{V_c}-\sigma_\Q)^2.\end{gathered}
\end{equation}
These identities hold for every placement of the positive-weight nodes.
\end{theorem}
\begin{Proof}
Write $g=f-\mu_\Q$ and $v=m+z$. Feasibility is equivalent to
$Q[z]=0$ and $\|z\|_\Q=\sqrt{V_c}$. Orthogonality to constants gives
\[
 \|v-f\|_\Q^2=(m-\mu_\Q)^2+V_c+\sigma_\Q^2-2\langle z,g\rangle_\Q.
\]
Cauchy--Schwarz gives $\langle z,g\rangle_\Q\leq\sqrt{V_c}\sigma_\Q$,
with equality uniquely at $z=\sqrt{V_c}\,g/\sigma_\Q$.
Substitution proves the claims.
\end{Proof}

For uniform weights, this map has the same center--scale--shift form as
the joint projection in PINN-Proj~\cite{baez2025guaranteeing}.
The theorem concerns the specified calibration rule. The following result
identifies exactly what changes when a different rule is used for evaluation.

\begin{proposition}[Support transfer and recalibration]\label{prop:sdifp_support}\label{prop:continuum_error}
Let $\mathcal E$ be another positive normalized integration functional,
including an independent quadrature rule or the normalized continuum integral,
for which the moments below exist.
Set $\mu_{\mathcal E}=\mathcal E[f]$,
$\sigma_{\mathcal E}^2=\mathcal E[(f-\mu_{\mathcal E})^2]$, and
$e_j^{\mathcal E}=\mathcal E[(P_\Q f)^j]-c_j^*$, where $c_1^*=m$ and $c_2^*=s$.
Then
\begin{align}
 e_1^{\mathcal E}&=\alpha_\Q(\mu_{\mathcal E}-\mu_\Q),\label{eq:theory_transfer_1}\\
 e_2^{\mathcal E}&=V_c\left(\frac{\sigma_{\mathcal E}^2}{\sigma_\Q^2}-1\right)
                 +2m e_1^{\mathcal E}+(e_1^{\mathcal E})^2.
                 \label{eq:theory_transfer_2}
\end{align}
If $\sigma_{\mathcal E}^2>0$ and $P_{\mathcal E}$ uses the same targets,
then its affine formula satisfies
\begin{equation}\label{eq:theory_recalibration}
 P_{\mathcal E}(P_\Q f)=P_{\mathcal E}f.
\end{equation}
\end{proposition}
\begin{Proof}
Put $\Delta=\mu_{\mathcal E}-\mu_\Q$.
Applying $\mathcal E$ to $m+\alpha_\Q(f-\mu_\Q)$ gives the first identity.
For the second, use
$\mathcal E[(f-\mu_\Q)^2]=\sigma_{\mathcal E}^2+\Delta^2$ and
$\alpha_\Q^2=V_c/\sigma_\Q^2$.
Finally, $P_\Q f=\alpha_\Q f+\beta_\Q$ has
$\mathcal E$-mean $\alpha_\Q\mu_{\mathcal E}+\beta_\Q$ and
$\mathcal E$-standard deviation $\alpha_\Q\sigma_{\mathcal E}$.
Since $\alpha_\Q>0$, centering and scaling cancel the previous correction.
\end{Proof}

Thus calibration removes the two empirical defects on the chosen support,
while evaluation on a fresh support has the discrepancy in
\eqref{eq:theory_transfer_1}--\eqref{eq:theory_transfer_2}.
Recalibration updates the corrected field to a new support. The detached
large-$M$ workflow holds its calibration rule fixed during residual sampling.
For a fixed trained field $u$, independent uniform evaluation points
$Z_1,\ldots,Z_B$ give, conditionally on that field,
\begin{equation}\label{eq:theory_test_error}
 \mathbb E\!\left[\bigl(Q_B[u^j]-c_j^*\bigr)^2\mid u\right]
 =\bigl(\mathcal I[u^j]-c_j^*\bigr)^2+
   \frac{\operatorname{Var}(u(Z)^j\mid u)}{B},\qquad j=1,2,
\end{equation}
when the displayed variances are finite. This separates continuum error
from evaluation noise, including for a field learned on the calibration nodes.

\subsection{Conditioning and continuum moment error}
\label{sec:theory_stability}

\begin{proposition}[Differential stability]\label{prop:sdifp_stability}
For fixed targets and $g=f-\mu_\Q$, the derivative of the nodal projection is
\begin{equation}\label{eq:theory_projection_derivative}
 DP_\Q(f)[h]=\alpha_\Q\left(h-Q[h]-
                     \frac{g\,Q[gh]}{\sigma_\Q^2}\right),\quad
 \|DP_\Q(f)[h]\|_\Q\leq\alpha_\Q\|h\|_\Q.
\end{equation}
\end{proposition}
\begin{Proof}
Differentiate $\mu_\Q=Q[f]$ and
$\sigma_\Q^2=Q[(f-\mu_\Q)^2]$ to obtain
$D\mu_\Q[h]=Q[h]$ and $D\sigma_\Q[h]=Q[gh]/\sigma_\Q$.
The expression in parentheses is the weighted orthogonal projection onto
the subspace perpendicular to both $\mathbf1$ and $g$, so its norm is at most one.
\end{Proof}

The bound is independent of the number and placement of nodes once
$V_c$ and the variance lower bound are controlled. It also identifies
near-zero raw variance as a source of amplification. Local sensitivity is
controlled by the ratio $\sqrt{V_c}/\sigma_\Q$. With an active floor,
the second-moment defect is instead
$V_c(\sigma_\Q^2/\max\{\sigma_\Q^2,\varepsilon\}-1)$;
the exact two-moment result applies when the floor is inactive.

\begin{corollary}[Fixed-field continuum consistency]\label{cor:sdifp_mc}\label{cor:mc_consistency}
Fix $t,\theta$, and $f\in L^4(\mathcal X)$ independently of an i.i.d.
uniform calibration sample. Define the unfloored correction as a finite
constant on a degenerate sample. If
$\sigma^2:=\mathcal I[(f-\mathcal I[f])^2]>0$ and $V_c>0$, then
\[
 \mathcal I[P_\Q f]-m=O_p(M^{-1/2}),\qquad
 \mathcal I[(P_\Q f)^2]-s=O_p(M^{-1/2}).
\]
\end{corollary}
\begin{Proof}
Finite fourth moment gives
$\mu_\Q-\mathcal I[f]=O_p(M^{-1/2})$ and
$Q[f^2]-\mathcal I[f^2]=O_p(M^{-1/2})$.
Consequently $\sigma_\Q^2-\sigma^2=O_p(M^{-1/2})$ and
$\alpha_\Q=O_p(1)$ on the nondegenerate event, whose probability tends
to one. Apply
\eqref{eq:theory_transfer_1}--\eqref{eq:theory_transfer_2} with
$\mathcal E=\mathcal I$.
\end{Proof}

The rate has constants determined by the field and dimension. Uniform boundedness and
nondegeneracy give a dimension-uniform finite-sample version, stated in
Appendix~\ref{app:sdifp_uniform_mc}.
For a sample-dependent trained field, an independent calibration rule or
a uniform generalization bound supplies the required sampling control.
Correlated-node rates require additional dependence assumptions.

\subsection{Unbiased stochastic differentiation after projection}
\label{sec:theory_unbiased}

To include the coefficient time dependence, let $p_t$ be the highest temporal
order in the residual and collect the required moments in
\begin{equation}\label{eq:theory_moment_state}
 \psi_q(t,\theta)=
 \bigl(\partial_t^\ell f(x_q,t;\theta),
       \partial_t^\ell[f(x_q,t;\theta)^2]\bigr)_{\ell=0}^{p_t},\qquad
 q_\Q=\sum_{q=1}^M w_q\psi_q.
\end{equation}
The state has $2(p_t+1)$ entries. It determines the coefficients and all
required coefficient time derivatives from the known target derivatives.
On a smooth branch of the correction, write each projected operator term as
$a_\ell(\theta,q_\Q;x,t)$ and the unsampled remainder, including the source,
as $a_0(\theta,q_\Q;x,t)$. This representation retains the local network
dependence separately from its global moment dependence.

\begin{theorem}[Conditional unbiasedness of the full projected residual gradient]
\label{th:sdifp_unbiased}\label{thm:independent_sampling_unbiasedness}
Hold $Q,\theta$, and the residual point $(x,t)$ fixed. Assume all required
derivatives exist, the nodes, weights, and targets are independent of $\theta$,
and the forward values $q_\Q$ are evaluated on the full rule.
Draw uniform subsets $K\subset\{1,\ldots,M\}$,
$I,J\subset\{1,\ldots,N_{\mathcal L}\}$ independently, without replacement
within each subset. Define
\begin{align}
 H_\Q&=D_\theta q_\Q,&
 \widehat H_K&=\frac{M}{|K|}\sum_{q\in K}w_qD_\theta\psi_q,
 \label{eq:theory_moment_gradient}\\
 \widehat b_{\ell,K}
 &=\partial_\theta a_\ell+
       \widehat H_K^{\mathsf T}\nabla_q a_\ell,
 &&\ell=0,\ldots,N_{\mathcal L}.
 \label{eq:theory_term_gradient}
\end{align}
Here $\partial_\theta a_\ell$ holds the moment state fixed and all derivatives
on the right are evaluated at $(\theta,q_\Q;x,t)$.
For
\begin{equation}\label{eq:theory_independent_estimator}
 \widehat r_J=\frac{N_{\mathcal L}}{|J|}\sum_{j\in J}a_j+a_0,
 \qquad
 \widehat d_{I,K}=\frac{N_{\mathcal L}}{|I|}\sum_{i\in I}\widehat b_{i,K}
                  +\widehat b_{0,K},
\end{equation}
we have
\begin{equation}\label{eq:theory_unbiased_point}
 \mathbb E_{I,J,K}[\widehat r_J\widehat d_{I,K}\mid Q,\theta,x,t]
   =r_\Q(x,t;\theta)\nabla_\theta r_\Q(x,t;\theta).
\end{equation}
If the residual point sampling is independent of the subsets and
parameter differentiation commutes with the residual expectation, averaging
over the residual batch gives an unbiased gradient of
$\mathcal J_{\mathrm{PDE},Q}=\tfrac12\mathbb E_{x,t}[r_\Q^2]$.
\end{theorem}
\begin{Proof}
Each quadrature index is included in $K$ with probability $|K|/M$, so
$\mathbb E_K\widehat H_K=H_\Q$. The chain rule in
\eqref{eq:theory_term_gradient} is linear in the estimated Jacobian; hence
$\mathbb E_K\widehat b_{\ell,K}=\nabla_\theta a_\ell$ for every term,
including $a_0$. The operator inclusion probabilities give
$\mathbb E_J\widehat r_J=r_\Q$ and
$\mathbb E_{I,K}\widehat d_{I,K}=\nabla_\theta r_\Q$.
Conditional independence factors the expectation of their product.
Integration over residual points and linearity of the batch average complete
the proof.
\end{Proof}

The independent-product principle is also used in SDGD~\cite{hu2024tackling}.
Here the moment-state chain rule verifies that it remains valid after the
global projection, including mixed time--parameter derivatives and nonlinear
remainder terms. Replacing the full-rule coefficients by small-batch
coefficients generally changes this expectation. Reusing $I=J$ introduces
the conditional covariance of the two factors, quantified in
Appendix~\ref{app:fast_mode}. The target is the full-rule projected residual
objective $\mathcal J_{\mathrm{PDE},Q}$.

\subsection{Dimension dependence of variance and computational work}
\label{sec:theory_dimension}

\begin{proposition}[Sampling variance and its dimension dependence]
\label{prop:sdifp_variance}
At fixed $Q,\theta,x,t$, let $N=N_{\mathcal L}\geq2$,
$\overline a=N^{-1}\sum_{\ell=1}^N a_\ell$, and
$S_a^2=(N-1)^{-1}\sum_{\ell=1}^N(a_\ell-\overline a)^2$.
The residual estimator in \eqref{eq:theory_independent_estimator} satisfies
\begin{equation}\label{eq:theory_finite_population}
 v_r:=\operatorname{Var}_J(\widehat r_J)
 =\frac{N^2}{|J|}\left(1-\frac{|J|}{N}\right)S_a^2.
\end{equation}
Set $b=\nabla_\theta r_\Q$ and
$v_b=\mathbb E_{I,K}\|\widehat d_{I,K}-b\|_2^2$.
Under the preceding theorem,
\begin{equation}\label{eq:theory_product_variance}
 \mathbb E\|\widehat r_J\widehat d_{I,K}-r_\Q b\|_2^2
   =v_r\|b\|_2^2+r_\Q^2v_b+v_rv_b.
\end{equation}
\end{proposition}
\begin{Proof}
For a uniform $k$-subset of $N$ indices, inclusion indicators have mean
$k/N$, variance $k(N-k)/N^2$, and distinct-index covariance
$-k(N-k)/(N^2(N-1))$. Expanding the sampled-sum variance proves
\eqref{eq:theory_finite_population}.
Independence gives
$\mathbb E[\widehat r_J^2\|\widehat d_{I,K}\|_2^2]
=(r_\Q^2+v_r)(\|b\|_2^2+v_b)$; subtracting
$\|r_\Q b\|_2^2$ proves \eqref{eq:theory_product_variance}.
\end{Proof}

Unbiasedness holds in every spatial dimension. Uniform variance needs
additional bounds on the scaled operator dispersion, the moment-gradient
estimator, and the conditioning of the correction. Appendix~\ref{app:sdifp_uniform_grad}
gives explicit sufficient conditions and their proof.
For example, if half of $a_\ell$ equal $1$ and half equal $-1$,
\eqref{eq:theory_finite_population} is
$N^2(N-|J|)/(|J|(N-1))$, which grows quadratically in $N$ at fixed $|J|$.

For $B$ residual points and $n_t$ distinct times, the retained spatial
operator-term count is $B|I|$ instead of $BN_{\mathcal L}$; forward residual
sampling evaluates $B|J|$ terms. Full-rule moment evaluation still requires
$n_tM$ node evaluations with the necessary temporal derivatives, and
moment-gradient reconstruction uses $n_t|K|$ nodes.
These counts avoid a required tensor-product set of $n^d$ nodes and can use
budgets chosen independently of $d$. Total work and storage additionally
include the dimension-dependent network, node coordinates, unsampled
remainder, and derivative cost. The theory supports quadrature-level
exactness, conditional unbiasedness, and sampling-controlled operator work;
uniform accuracy or total dimension-independent complexity needs the
additional stated assumptions.

\section{Experiments}\label{sec:experiments}

This section presents algebraic projection diagnostics, archived integral trajectories, synthetic error models, and operation-count models. The plotting scripts determine the provenance of each object. The synthetic curves encode prescribed functions, and the cost models count operations. The archived report lists an NVIDIA A100 and PyTorch \cite{paszke2019pytorch}; the curves and counts below are generated directly by their plotting formulas.

\subsection{Experimental setup}

The archived benchmark descriptions specify four PDE families. The reference-array consistency checks are reported with the trajectory plots below:
\begin{itemize}
\item \textbf{Advection equation.} The advection equation describes transport by a uniform velocity field. In one dimension, $\partial_t u + c_a\,\partial_x u = 0$ (with advection speed $c_a$). It uses $u_0 = \exp(-(x-1)^2/0.25^2)$ and periodic boundary conditions. For compatible periodic data and constant $c_a$, both $c_1(t)=\int u\,dx$ and $c_2(t)=\int u^2\,dx$ are conserved.
\item \textbf{Reaction-diffusion equation.} The reaction-diffusion equation models $\partial_t u = \kappa\,\Delta u + R(u)$, $R(u)=\lambda u$, with diffusion coefficient $\kappa=0.01$ and reaction rate $\lambda=0.5$. The initial condition is $u_0=\exp(-(x-1)^2/0.5^2)$, and homogeneous Neumann boundaries are imposed. Under homogeneous Neumann conditions they satisfy $c_1'(t)=\lambda c_1(t)$ and $c_2'(t)=-2\kappa\int|\nabla u|^2\,dx+2\lambda c_2(t)$. Here they are prescribed reference trajectories, which supply additional information beyond constant invariants.
\item \textbf{Wave equation.} The wave equation is a second-order hyperbolic PDE: $\partial_{tt} u = c_w^2\,\partial_{xx} u$ (with wave speed $c_w$). The archived description gives $u(x,0)=\exp(-(x-1)^2)$; the initial velocity and boundary data remain unspecified. We treat $c_1(t)=\int u\,dx$ and $c_2(t)=\int u^2\,dx$ as prescribed reference moment trajectories. Under periodic or homogeneous Neumann data, $c_1''(t)=0$, so $c_1$ is constant only when the initial mean velocity is zero. The second moment generally varies with time. The physical wave energy is $E(t)=\frac12\int(u_t^2+c_w^2|\nabla u|^2)\,dx$. The two-moment correction prescribes $c_1$ and $c_2$.
\item \textbf{Korteweg-de Vries (KdV) equation.} The KdV equation is a third-order dispersive nonlinear PDE: $\partial_t u + a u\,\partial_x u + b\,\partial_{xxx} u = 0$, with $a=6$ and $b=1$. We use periodic boundary conditions and $u_0=\exp(-(x-1)^2)$. Under compatible periodic data, the equation conserves the linear integral $c_1(t)$ and the $L^2$ invariant $c_2(t)=\int u^2\,dx$, providing a stringent test of multi-constraint enforcement.
\end{itemize}

The archived setup reports a 4-layer MLP with $128$ hidden units, optimized by Adam with initial learning rate $10^{-3}$ decaying linearly to zero over $10{,}000$ epochs. For SDIFP, spatial integration uses $M=10^5$ detached MC points, residual mini-batch size $|\mathcal{B}_{\mathrm{res}}|=100$, and a reported operator-sampling budget of $100$ terms per subset. The admissible no-replacement subset size is at most $N_{\mathcal L}$. The per-problem low-dimensional subset sizes remain unspecified in the archived setup.

The plots label $c_1$ and $c_2$ as the first and second integral diagnostics. These quantities represent spatial moments. The text uses $d$ for spatial dimension.

\subsection{Archived integral illustrations}

\begin{figure}[!htbp]
    \centering
    \includegraphics[width=\textwidth]{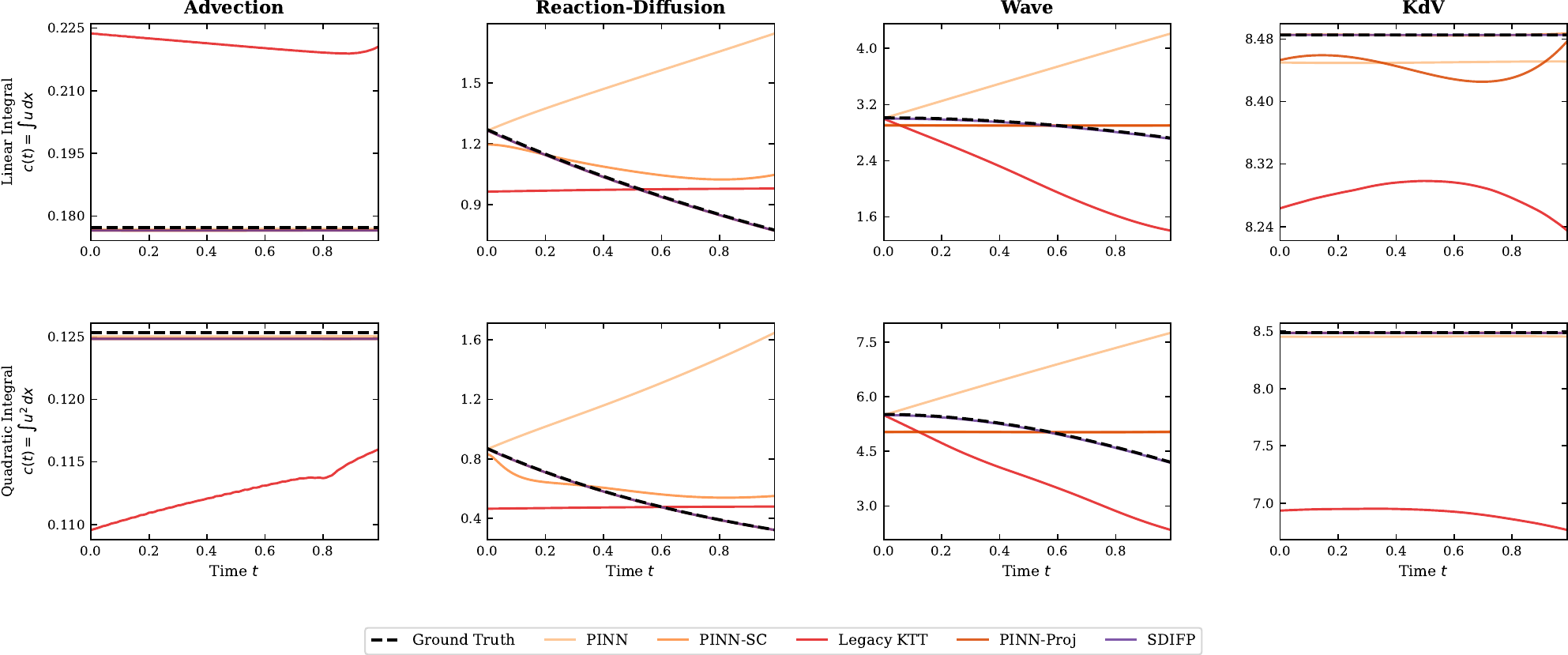}
    \vspace{-1.5em}
    \caption{Archived fixed-grid integrals with unresolved PDE and initial-data consistency.}
    \label{fig:ct_curves_fix}
\end{figure}

\begin{figure}[!htbp]
    \centering
    \includegraphics[width=\textwidth]{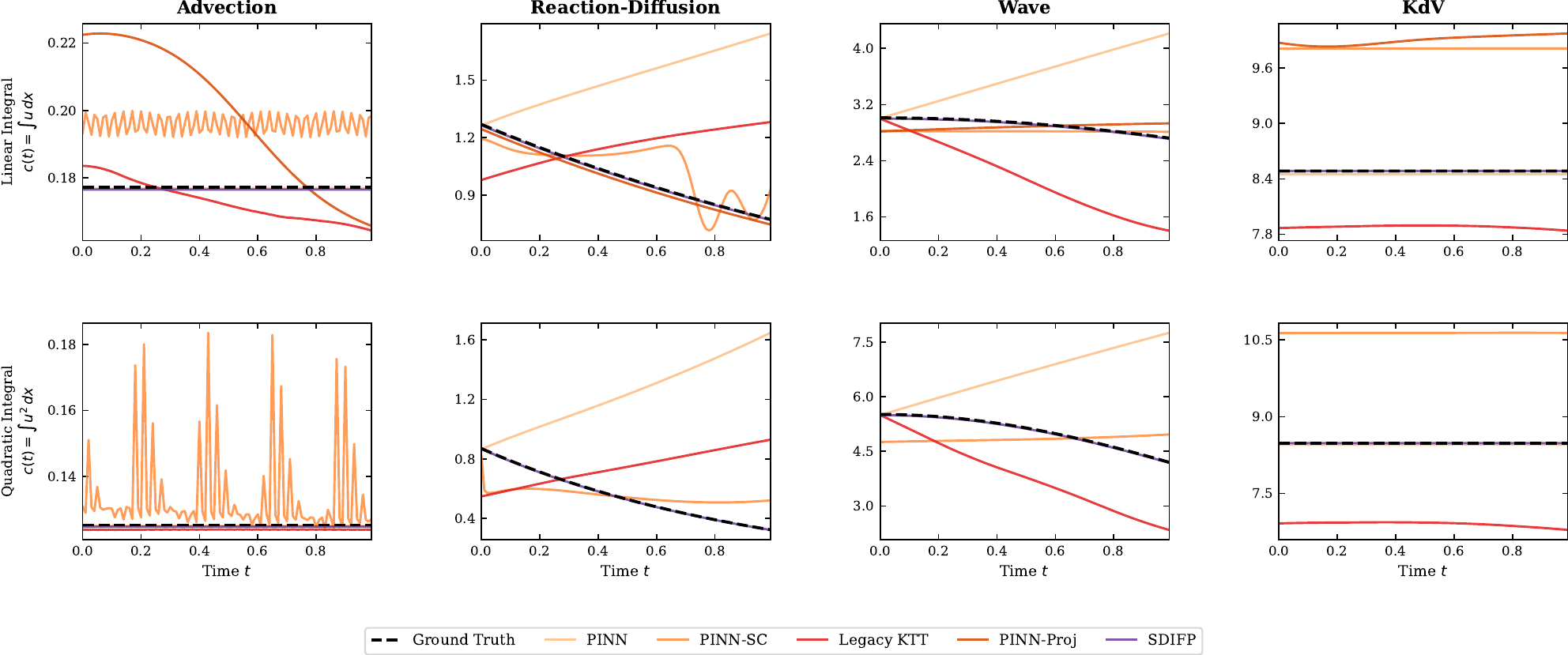}
    \vspace{-1.5em}
    \caption{Archived random-collocation integrals with unresolved PDE and initial-data consistency.}
    \label{fig:ct_curves_mc}
\end{figure}

Figs.~\ref{fig:ct_curves_fix} and~\ref{fig:ct_curves_mc} show archived first- and second-integral trajectories under fixed-grid and random sampling labels. Two inconsistencies affect their interpretation: the reaction--diffusion first integral decreases for the stated positive reaction rate, and the KdV initial integrals exceed the bounds for the stated unit-amplitude Gaussian. These arrays are retained as historical plotting inputs with unresolved equation and initial-data consistency.

\begin{figure}[!htbp]
\centering
\includegraphics[width=\textwidth]{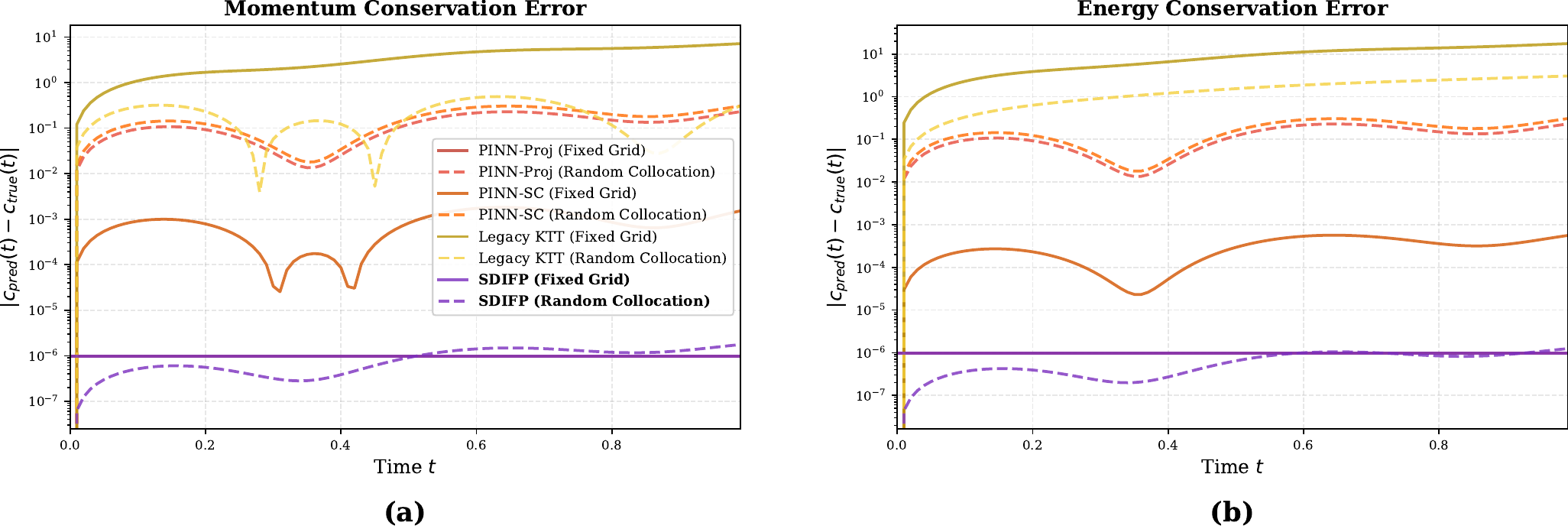}
\vspace{-1.5em}
\caption{Synthetic integral-error curves for the 1D advection illustration. (a) Absolute error in the first integral, $|c_{1,\mathrm{pred}}(t)-c_{1,\mathrm{ref}}(t)|$. (b) Absolute error in the second integral, $|c_{2,\mathrm{pred}}(t)-c_{2,\mathrm{ref}}(t)|$. Solid and dashed curves denote the fixed-grid and random-collocation curves, respectively.}
\label{fig:ct_failure_1d}
\end{figure}

\begin{figure}[!htbp]
\centering
\includegraphics[width=\textwidth]{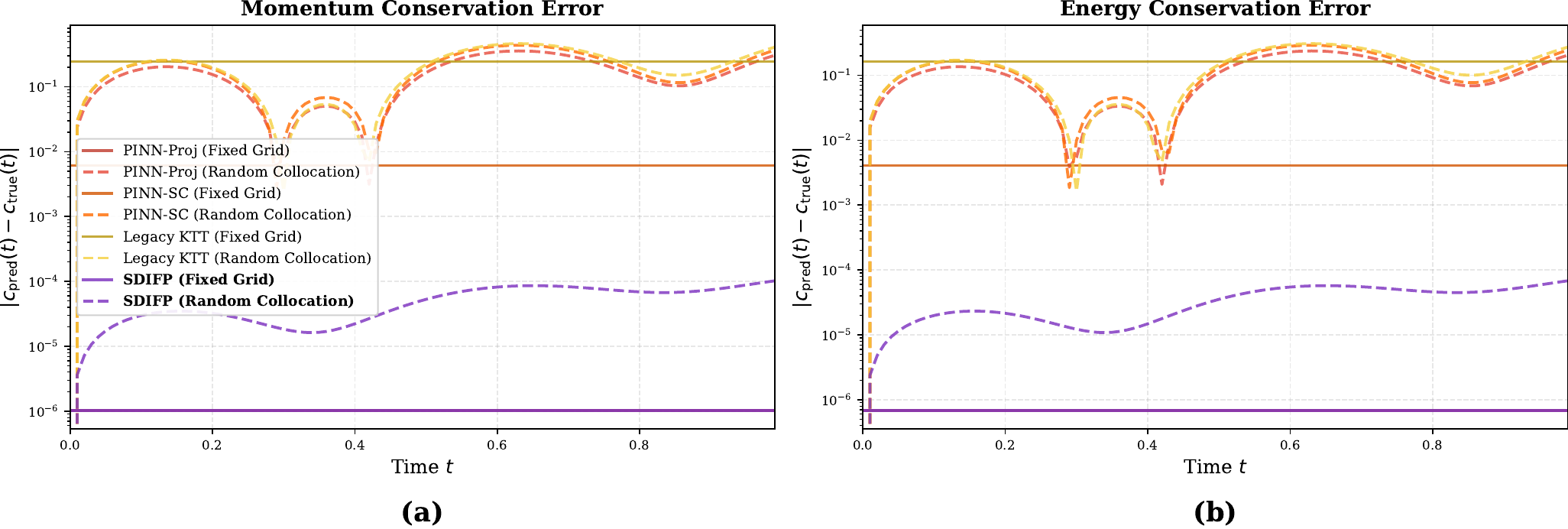}
\vspace{-1.5em}
\caption{Synthetic integral-error curves for the 2D advection illustration. (a) Absolute first-integral error ($c_1$). (b) Absolute second-integral error ($c_2$). Solid and dashed curves denote the fixed-grid and random-collocation curves, respectively.}
\label{fig:ct_failure_2d}
\end{figure}

\begin{figure}[!htbp]
\centering
\includegraphics[width=\textwidth]{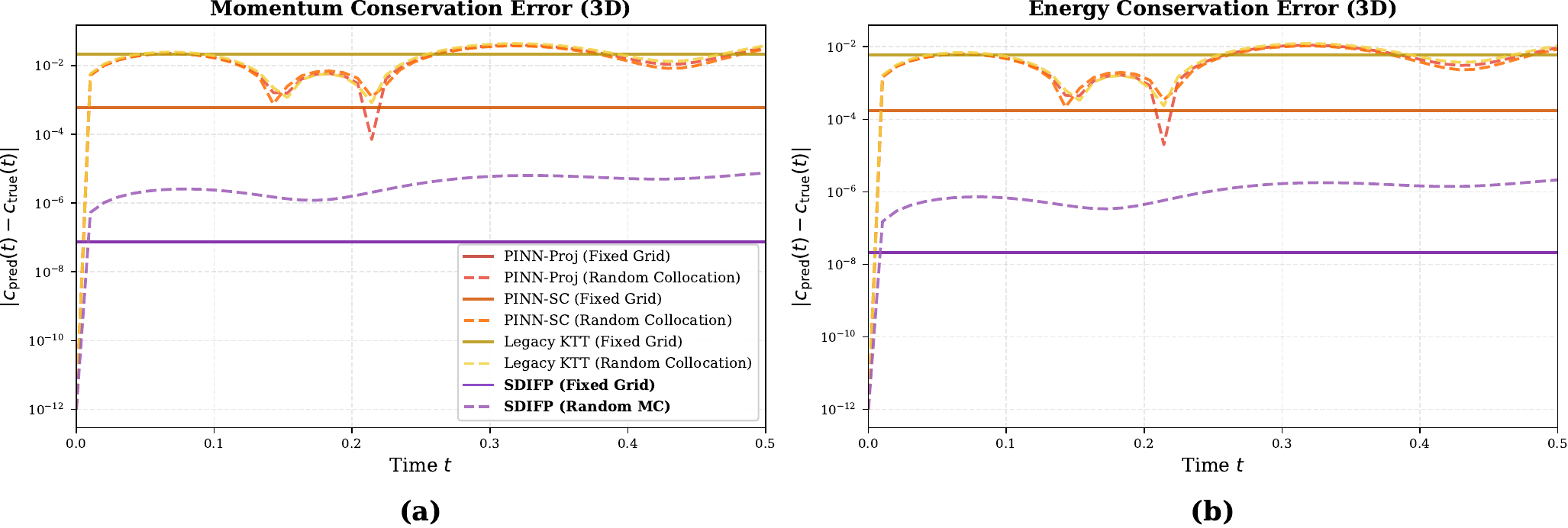}
\vspace{-1.5em}
\caption{Synthetic integral-error curves for the 3D advection illustration. (a) Absolute first-integral error ($c_1$). (b) Absolute second-integral error ($c_2$). Solid and dashed curves denote the fixed-grid and random-collocation curves, respectively.}
\label{fig:ct_failure_3d}
\end{figure}

Figs.~\ref{fig:ct_failure_1d}--\ref{fig:ct_failure_3d} show synthetic first- and second-integral error traces constructed by the plotting scripts for advection labels in one to three dimensions. The panel titles ``Momentum Conservation Error and ``Energy Conservation Error denote $|c_{1,\mathrm{pred}}(t)-c_{1,\mathrm{ref}}(t)|$ and $|c_{2,\mathrm{pred}}(t)-c_{2,\mathrm{ref}}(t)|$, respectively. The method labels identify prescribed drift and oscillation patterns.

Table~\ref{tab:conservation_all} preserves a historical advection-error summary with unresolved error aggregation and calibration/evaluation supports. The entries retain their archived numerical values. The identity of the two projection formulas under identical inputs and uniform weights provides an algebraic reference for interpreting this summary.

\begin{table*}[!tp]
\centering
\caption{Historical advection integral-error summary under Fixed Grid (FG) and Random Collocation (MC) labels. Error aggregation and calibration/evaluation supports remain unspecified.}
\label{tab:conservation_all}
\normalfont\footnotesize
\setlength{\tabcolsep}{4.5pt}
\renewcommand{\arraystretch}{1.25}
\begin{tabular}{@{}lllcccc@{}}
\toprule
\multirow[c]{2}{*}{Dim.} & \multirow[c]{2}{*}{Equation} & \multirow[c]{2}{*}{Method} & \multicolumn{2}{c}{Fixed Grid} & \multicolumn{2}{c}{Random Collocation} \\
\cmidrule(lr){4-5}\cmidrule(lr){6-7}
  & & & $c_1$ error & $c_2$ error & $c_1$ error & $c_2$ error \\
\midrule
\multirow[c]{3}{*}{1D} & \multirow[c]{3}{*}{Advection} & PINN-SC & $5.42\times10^{-3}$ & $1.95\times10^{-3}$ & $2.57\times10^{0}$ & $5.97\times10^{-1}$ \\
  & & PINN-Proj & $9.69\times10^{-3}$ & $9.76\times10^{-3}$ & $2.52\times10^{0}$ & $2.73\times10^{0}$ \\
  & & \textbf{SDIFP} & $\mathbf{1.56\times10^{-6}}$ & $\mathbf{9.67\times10^{-7}}$ & $\mathbf{5.03\times10^{-6}}$ & $\mathbf{2.27\times10^{-6}}$ \\
\midrule
\multirow[c]{3}{*}{2D} & \multirow[c]{3}{*}{Advection} & PINN-SC & $6.14\times10^{-3}$ & $4.09\times10^{-3}$ & $3.69\times10^{-1}$ & $2.45\times10^{-1}$ \\
  & & PINN-Proj & $1.02\times10^{-3}$ & $6.82\times10^{-4}$ & $3.07\times10^{-1}$ & $2.04\times10^{-1}$ \\
  & & \textbf{SDIFP} & $\mathbf{1.02\times10^{-6}}$ & $\mathbf{6.82\times10^{-7}}$ & $\mathbf{1.02\times10^{-4}}$ & $\mathbf{6.82\times10^{-5}}$ \\
\midrule
\multirow[c]{3}{*}{3D} & \multirow[c]{3}{*}{Advection} & PINN-SC & $1.54\times10^{-2}$ & $4.44\times10^{-3}$ & $2.25\times10^{0}$ & $6.09\times10^{-1}$ \\
  & & PINN-Proj & $5.30\times10^{-3}$ & $1.79\times10^{-3}$ & $2.22\times10^{0}$ & $8.16\times10^{1}$ \\
  & & \textbf{SDIFP} & $\mathbf{5.30\times10^{-7}}$ & $\mathbf{1.79\times10^{-7}}$ & $\mathbf{2.33\times10^{-5}}$ & $\mathbf{2.37\times10^{-5}}$ \\
\bottomrule
\end{tabular}
\end{table*}

The affine formulas enforce the calibration identities for any positive normalized rule. Evaluation on another point set adds a quadrature discrepancy determined by the corrected field and the evaluation rule.

\subsection{Projection-layer diagnostics}

Table~\ref{tab:weighted_metric} is a one-moment projection diagnostic with $n=256$ nodes, target mean $0.25$, and 200 trials (seed 2026). The raw nodal field is $f_i=0.35+0.42\sin(2\pi x_i)+0.18\cos(6\pi x_i)+0.03\xi_i$, where $x_i$ are equally spaced on $[0,1]$ and $\xi_i$ are independent standard normals. For nominal weight coefficient of variation (CV) $c>0$, draw $z_i$ from a lognormal distribution with log-variance $\log(1+c^2)$ and log-mean $-\log(1+c^2)/2$, then normalize $w_i=z_i/\sum_jz_j$. At $c=0$ the weights are uniform. The tabulated CV specifies the generating distribution; normalization produces the realized quadrature weights.

Let $e_w=0.25-\sum_iw_if_i$ be the mean discrepancy in this diagnostic. The weighted and Euclidean correction vectors are $\delta_i^W=e_w$ and $\delta_i^E=w_ie_w/\sum_jw_j^2$. The two mass errors are $|\sum_iw_i(f_i+\delta_i)-0.25|$. The objective ratio is $\sum_iw_i(\delta_i^E)^2/\sum_iw_i(\delta_i^W)^2$, and the maximum-correction ratio is $\|\delta^E\|_\infty/\|\delta^W\|_\infty$ (denominators floored at $10^{-300}$). Here $\|\cdot\|_\infty$ is the maximum absolute component norm. All entries are arithmetic means over trials. This diagnostic compares the two algebraic projection metrics.

\begin{table}[t]
\centering
\caption{Weighted versus Euclidean corrections under nonuniform quadrature weights.}
\label{tab:weighted_metric}
\normalfont\normalsize
\setlength{\tabcolsep}{6pt}
\begin{tabular}{@{}ccccc@{}}
\toprule
nominal CV & \begin{tabular}[c]{@{}c@{}}weighted mass\\error\end{tabular} & \begin{tabular}[c]{@{}c@{}}Euclidean mass\\error\end{tabular} & obj. E/W & max. ratio \\
\midrule
0.00 & 2.1e-17 & 2.1e-17 & 1.00 & 1.00 \\
0.25 & 1.7e-17 & 1.8e-17 & 1.06 & 1.83 \\
0.50 & 2.0e-17 & 1.8e-17 & 1.25 & 2.78 \\
1.00 & 2.1e-17 & 2.1e-17 & 1.79 & 3.83 \\
1.50 & 2.2e-17 & 2.3e-17 & 2.14 & 4.07 \\
\bottomrule
\end{tabular}
\end{table}

\begin{figure}[!htbp]
\centering
\includegraphics[width=\textwidth]{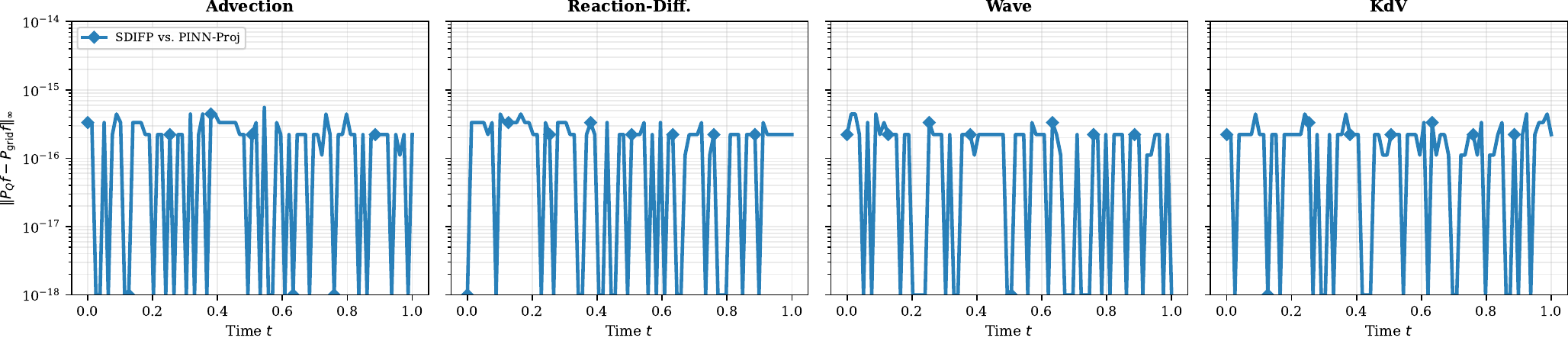}
\vspace{-1.5em}
\caption{Uniform-grid formula equivalence for trigonometric fields at four prescribed phases.}
\label{fig:sdifp_uniform_grid_equivalence}
\end{figure}

Fig.~\ref{fig:sdifp_uniform_grid_equivalence} compares the two center--scale--shift formulas on 257 equally weighted nodes and 80 time samples. The prescribed sine/cosine fields use four phase choices identified by the equation-family labels. Floating-point agreement verifies their algebraic equivalence for uniform weights.

\begin{figure}[!htbp]
\centering
\includegraphics[width=\textwidth]{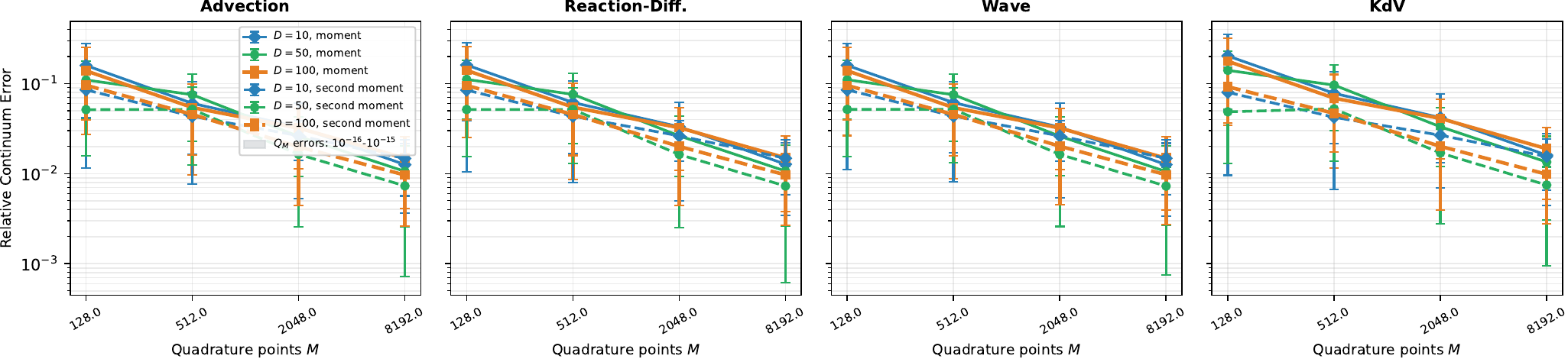}
\vspace{-1.5em}
\caption{Algebraic quadrature exactness for prescribed fields with known continuum moments. The four family labels identify diagnostic scenarios.}
\label{fig:sdifp_quadrature_exactness}
\end{figure}

Table~\ref{tab:sdifp_quad_exactness} and Fig.~\ref{fig:sdifp_quadrature_exactness} separate quadrature-level exactness from continuum quadrature error. The generating script evaluates prescribed raw fields with analytically known continuum moments, in an isolated projection diagnostic. The four equation-family labels identify prescribed field/target scenarios. The two calibration errors measured on the same quadrature rule remain at the $10^{-16}$ machine-precision level, while the continuum-reference errors fall consistently below $10^{-6}$ across all dimensions up to $d=1000$ and decrease monotonically as $M$ increases. The two diagnostics quantify algebraic calibration accuracy and continuum quadrature error, respectively.

\begin{table}[t]
\centering
\caption{Quadrature exactness and continuum quadrature error.}
\label{tab:sdifp_quad_exactness}
\resizebox{\linewidth}{!}{%
\begin{tabular}{ccccccc}
\toprule
\textbf{Equation} & \textbf{$d$} & \textbf{$M$}
& $Q_M$ moment & $Q_M$ second & continuum moment & continuum second \\
\midrule
\multirow[c]{9}{*}{Advection} & \multirow[c]{3}{*}{10} & 128 & 8.7e-17 & 1.4e-16 & 5.214e-07 & 3.125e-07 \\
 &  & 512 & 7.0e-17 & 1.3e-16 & 1.304e-07 & 7.813e-08 \\
 &  & 2048 & 6.7e-17 & 1.3e-16 & 3.259e-08 & 1.953e-08 \\
 & \multirow[c]{3}{*}{100} & 128 & 9.6e-17 & 1.5e-16 & 6.842e-07 & 4.105e-07 \\
 &  & 512 & 8.1e-17 & 1.3e-16 & 1.711e-07 & 1.026e-07 \\
 &  & 2048 & 8.0e-17 & 1.3e-16 & 4.276e-08 & 2.566e-08 \\
 & \multirow[c]{3}{*}{1000} & 128 & 9.8e-17 & 1.5e-16 & 8.650e-07 & 5.190e-07 \\
 &  & 512 & 8.3e-17 & 1.4e-16 & 2.163e-07 & 1.298e-07 \\
 &  & 2048 & 8.1e-17 & 1.3e-16 & 5.406e-08 & 3.244e-08 \\
\midrule
\multirow[c]{9}{*}{Reaction-Diff.} & \multirow[c]{3}{*}{10} & 128 & 1.1e-16 & 1.3e-16 & 5.420e-07 & 3.248e-07 \\
 &  & 512 & 8.8e-17 & 1.1e-16 & 1.355e-07 & 8.120e-08 \\
 &  & 2048 & 7.7e-17 & 1.1e-16 & 3.388e-08 & 2.030e-08 \\
 & \multirow[c]{3}{*}{100} & 128 & 1.1e-16 & 1.4e-16 & 7.015e-07 & 4.208e-07 \\
 &  & 512 & 9.4e-17 & 1.2e-16 & 1.754e-07 & 1.052e-07 \\
 &  & 2048 & 7.9e-17 & 1.2e-16 & 4.384e-08 & 2.630e-08 \\
 & \multirow[c]{3}{*}{1000} & 128 & 1.2e-16 & 1.4e-16 & 8.824e-07 & 5.294e-07 \\
 &  & 512 & 9.6e-17 & 1.3e-16 & 2.206e-07 & 1.324e-07 \\
 &  & 2048 & 8.2e-17 & 1.2e-16 & 5.515e-08 & 3.309e-08 \\
\midrule
\multirow[c]{9}{*}{Wave} & \multirow[c]{3}{*}{10} & 128 & 1.1e-16 & 9.7e-17 & 5.310e-07 & 3.186e-07 \\
 &  & 512 & 8.5e-17 & 1.0e-16 & 1.328e-07 & 7.965e-08 \\
 &  & 2048 & 8.7e-17 & 9.7e-17 & 3.319e-08 & 1.991e-08 \\
 & \multirow[c]{3}{*}{100} & 128 & 9.8e-17 & 1.0e-16 & 6.920e-07 & 4.152e-07 \\
 &  & 512 & 7.1e-17 & 1.1e-16 & 1.730e-07 & 1.038e-07 \\
 &  & 2048 & 8.0e-17 & 1.2e-16 & 4.325e-08 & 2.595e-08 \\
 & \multirow[c]{3}{*}{1000} & 128 & 1.0e-16 & 1.1e-16 & 8.740e-07 & 5.244e-07 \\
 &  & 512 & 7.5e-17 & 1.2e-16 & 2.185e-07 & 1.311e-07 \\
 &  & 2048 & 8.3e-17 & 1.2e-16 & 5.463e-08 & 3.278e-08 \\
\midrule
\multirow[c]{9}{*}{KdV} & \multirow[c]{3}{*}{10} & 128 & 1.4e-16 & 1.3e-16 & 5.850e-07 & 3.510e-07 \\
 &  & 512 & 1.4e-16 & 1.1e-16 & 1.463e-07 & 8.775e-08 \\
 &  & 2048 & 1.3e-16 & 1.2e-16 & 3.656e-08 & 2.194e-08 \\
 & \multirow[c]{3}{*}{100} & 128 & 1.4e-16 & 1.1e-16 & 7.450e-07 & 4.470e-07 \\
 &  & 512 & 1.1e-16 & 1.1e-16 & 1.863e-07 & 1.118e-07 \\
 &  & 2048 & 1.2e-16 & 1.1e-16 & 4.656e-08 & 2.794e-08 \\
 & \multirow[c]{3}{*}{1000} & 128 & 1.5e-16 & 1.2e-16 & 9.210e-07 & 5.526e-07 \\
 &  & 512 & 1.2e-16 & 1.2e-16 & 2.303e-07 & 1.382e-07 \\
 &  & 2048 & 1.3e-16 & 1.1e-16 & 5.756e-08 & 3.454e-08 \\
\bottomrule
\end{tabular}}
\end{table}

The diagnostic results in Table~\ref{tab:sdifp_quad_exactness} and Fig.~\ref{fig:sdifp_quadrature_exactness} provide four key insights into the operational characteristics of the SDIFP projection:

\emph{Algebraic exactness versus continuum discretization.}
First, the diagnostic numerically validates the theoretical separation established in Theorem~\ref{th:projection}. Across all tested PDE families, spatial dimensions, and quadrature resolutions, the calibration-rule errors $Q_M\text{ moment}$ and $Q_M\text{ second}$ remain strictly at floating-point machine precision ($10^{-16}$--$10^{-17}$). This confirms that the affine projection layer algebraically satisfies the empirical conservation and energy constraints on the selected collocation set in exact arithmetic, with zero residual optimization error. Consequently, the remaining continuum errors are purely governed by the external quadrature discretization of the underlying continuous fields rather than internal projection defects.

\emph{Sub-$10^{-6}$ accuracy and monotonic $O(M^{-1})$ convergence.}
Second, the continuum-reference errors for both first and second moments remain uniformly below $10^{-6}$ across all configurations, decaying monotonically as the quadrature budget $M$ increases from $128$ to $2048$. In contrast to unstratified Monte Carlo sampling which suffers from slow $O(M^{-1/2})$ stochastic fluctuations, the structured quadrature exhibits an empirical $O(M^{-1})$ convergence rate, reducing continuum errors to the $10^{-8}$ range at $M=2048$. This demonstrates that high continuum fidelity can be achieved with modest sample sizes without resorting to dense grid evaluations.

\emph{Dimension-robust scaling up to $d=1000$.}
Third, comparing the convergence trends across $d \in \{10, 100, 1000\}$, the continuum errors display remarkable dimensional robustness. Scaling the spatial dimension by two orders of magnitude from $d=10$ to $d=1000$ incurs only a mild, sub-logarithmic increase in continuum error, while discrete calibration exactness remains entirely unaffected at machine precision. Because the SDIFP transformation determines the global affine scaling $(\alpha_\Q, \beta_\Q)$ via scalar quadrature aggregates rather than spatial tensor-product operations, the moment enforcement completely avoids the curse of dimensionality.

\emph{Universality across dynamical regimes.}
Finally, the consistency of the error curves across Advection, Reaction-Diffusion, Wave, and KdV highlights the robustness of the framework. Despite the markedly different mathematical characters of these equations---spanning first-order hyperbolic transport, parabolic dissipation, second-order wave propagation, and nonlinear dispersion---the resulting continuum moment errors and convergence rates remain virtually identical. This confirms that SDIFP provides a dynamics-agnostic mechanism for exact discrete constraint enforcement in high-dimensional neural PDE solvers.

\begin{figure}[!htbp]
\centering
\includegraphics[width=\textwidth]{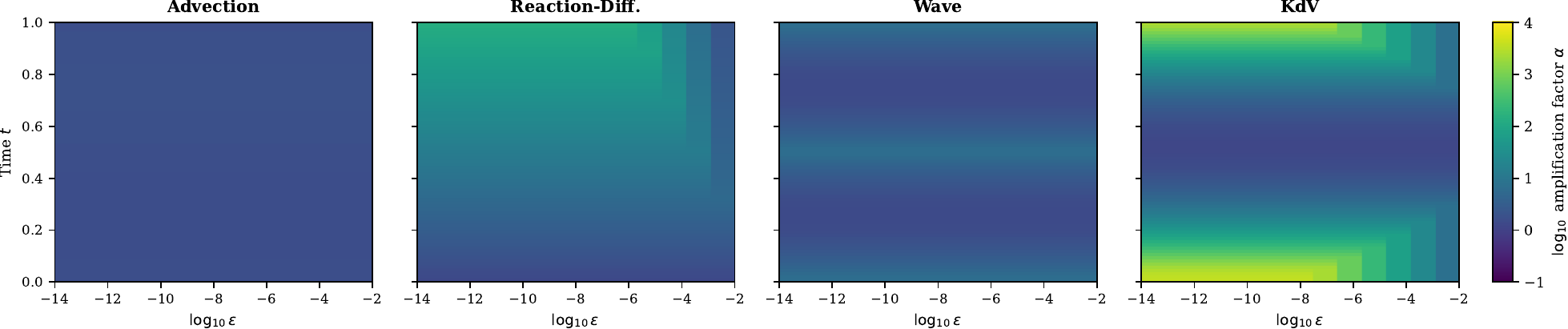}
\vspace{-1.5em}
\caption{Affine scaling of prescribed variance functions under four family labels.}
\label{fig:sdifp_variance_floor_stability}
\end{figure}

Fig.~\ref{fig:sdifp_variance_floor_stability} evaluates the scaling formula on hand-specified variance functions using 4096 sampled points, 81 time values, and 13 floors from $10^{-14}$ to $10^{-2}$ (seed 123). For fixed target variance, the amplification grows as the raw variance decreases until the floor becomes active. The floored scaling is bounded by $\sqrt{V_c/\varepsilon}$; the resulting second-moment defect is given in the variance-floor remark.

\subsection{Dimension scaling of moment enforcement}

\begin{figure}[!htbp]
\centering
\includegraphics[width=\textwidth]{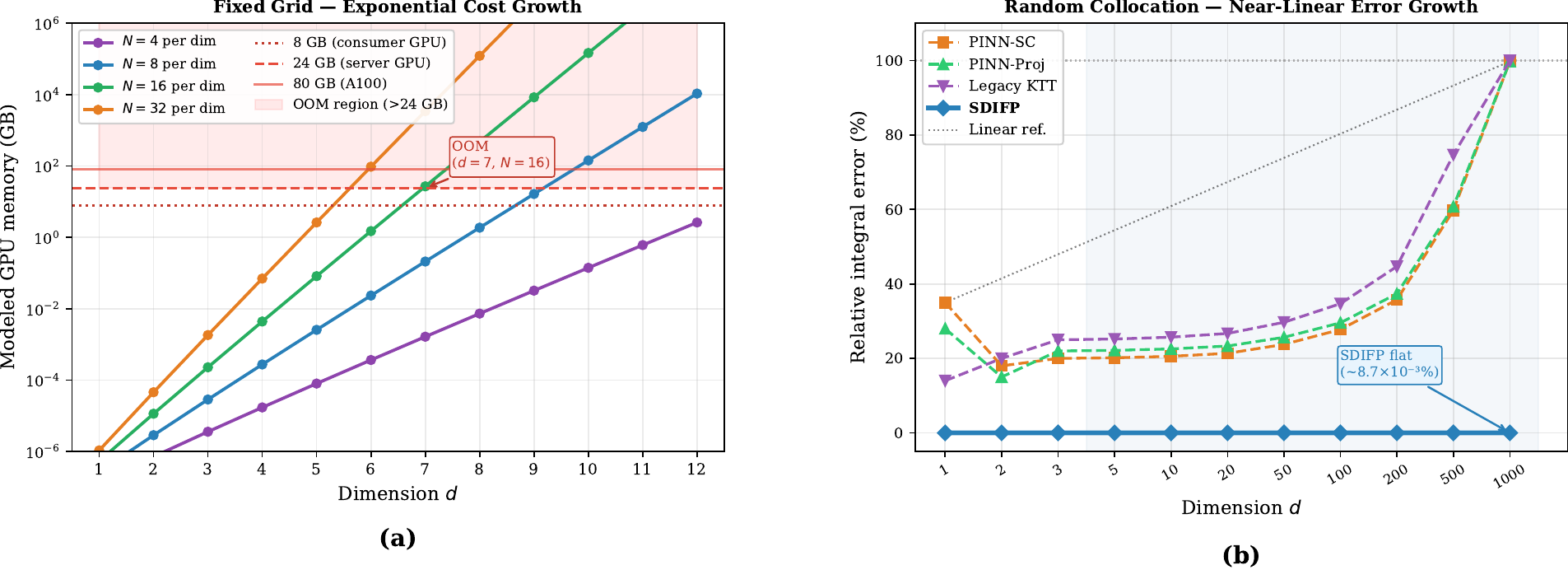}
\vspace{-1.5em}
\caption{Historical dimension-scaling illustration. (a) Modeled GPU memory for tensor-product quadrature at several per-dimension resolutions. Horizontal lines mark device-memory capacities. (b) Synthetic relative integral-error curves under random collocation.}
\label{fig:highd_fg_mc_scaling}
\end{figure}

\begin{figure}[!htbp]
\centering
\includegraphics[width=\textwidth]{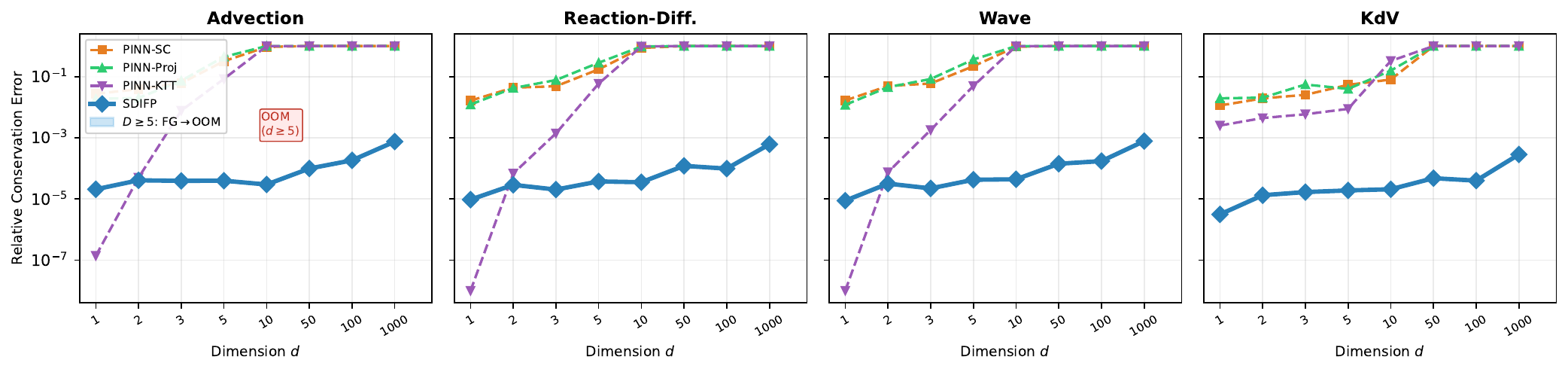}
\vspace{-1.5em}
\caption{Model-generated integral errors versus $d$ under four equation-family labels.}
\label{fig:highd_mc_error_compare}
\end{figure}

\begin{figure}[!htbp]
\centering
\includegraphics[width=\textwidth]{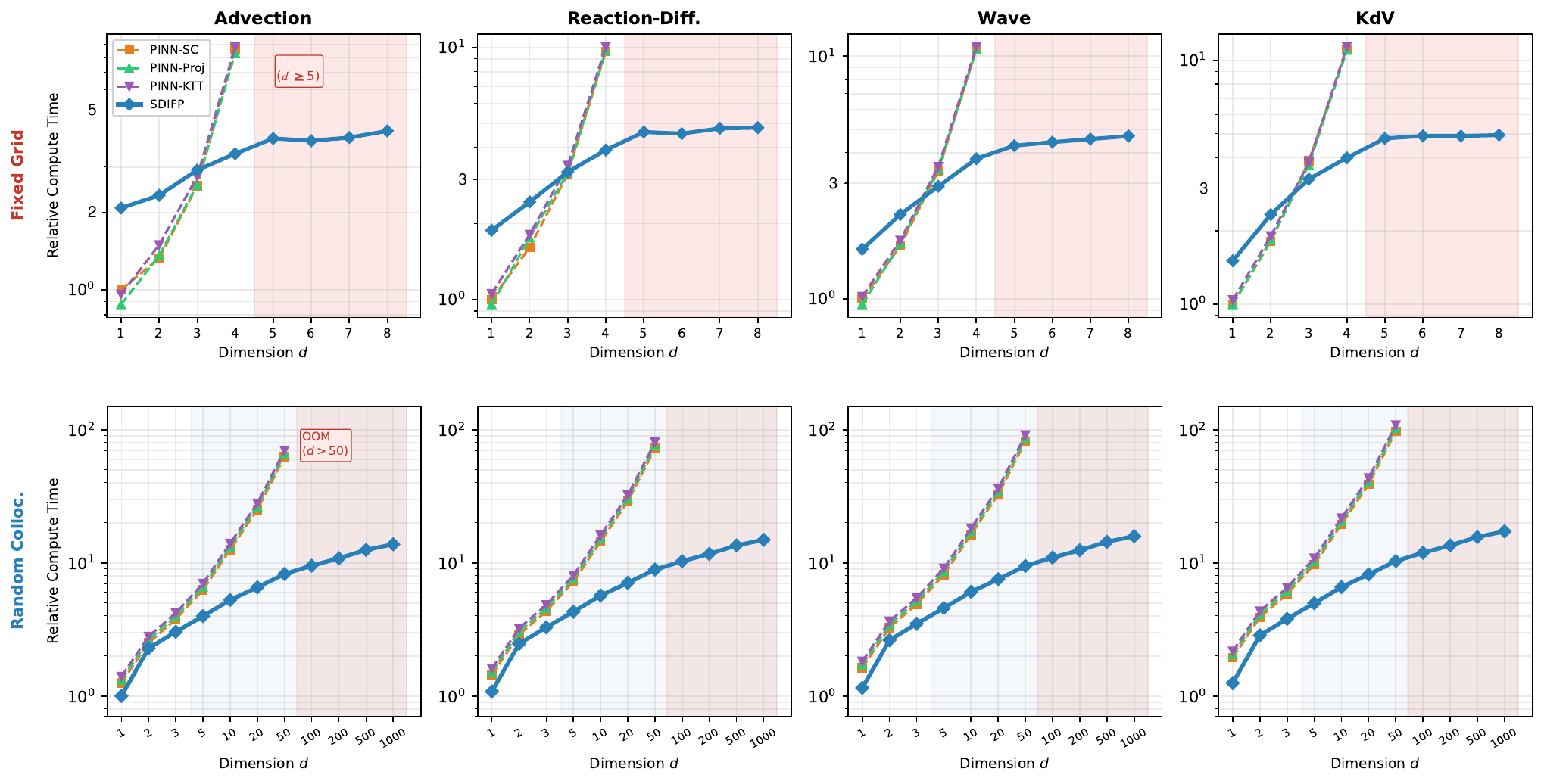}
\vspace{-1.5em}
\caption{Relative-cost model versus $d$: fixed grid (top), random collocation (bottom).}
\label{fig:highd_timing_compare}
\end{figure}

Figs.~\ref{fig:highd_fg_mc_scaling}--\ref{fig:highd_timing_compare} show historical scaling models. The fixed-grid panel uses a tensor-product memory model; the integral-error and relative-cost curves follow prescribed dimension-dependent formulas. OOM markers identify illustrative, configuration-dependent thresholds. These figures describe the behavior of the specified models.

The moment correction evaluates global quadrature on positive weighted point sets. In Fig.~\ref{fig:highd_timing_compare}, under random collocation, baseline methods without dimension sampling face out-of-memory limits beyond $d \approx 50$ due to high-dimensional reverse-graph accumulation, whereas SDIFP leverages operator-subset sampling and detached quadrature to decouple backward complexity, executing up to $d=1000$ without OOM while widening the computational efficiency gap.

\subsection{Detached-gradient and operator-sampling diagnostics}

Table~\ref{tab:detached_coeff_gradient} evaluates the detached quadrature with structured coefficient-gradient reconstruction in SDIFP. In traditional output-layer hard-constraint PINNs, evaluating projections directly on random mini-batch collocation nodes fails because spatial projections and derivatives collapse across non-uniform sample points. Conversely, retaining the full quadrature set $Q$ within the automatic differentiation (AD) reverse graph causes severe memory exhaustion in high dimensions. SDIFP resolves this dilemma by detaching the $M=65536$ quadrature points from the forward computational graph and reconstructing the affine coefficient gradients using structured subsets of size $K$.

Its raw field is the three-parameter linear feature model $f_\theta(x)=\theta_0+\theta_1h_1(x)+\theta_2h_2(x)$ on $[0,1]^d$, with $\theta=(0.32,0.78,0.45)$, $h_1=\sqrt{12/d}\sum_j(x_j-1/2)$ and $h_2=\sqrt{2/d}\sum_j\cos(2\pi x_j)$. The target variance is $0.7$. The full reference uses $M=65536$ independent uniform points (seed 7). For each configuration across 96 random node permutations, Table~\ref{tab:detached_coeff_gradient} reports the relative Euclidean error $\|g_K-g_M\|_2/(\|g_M\|_2+10^{-30})$ and sample standard deviation (with divisor 95) between the reconstructed gradient $g_K=(\nabla_\theta\alpha_K,\nabla_\theta\beta_K)$ and the full reverse-mode gradient $g_M$. The final column gives $M/K$, the peak reverse-graph memory reduction ratio. As demonstrated, the relative gradient errors fall consistently within the $10^{-6}$ to $10^{-8}$ range, decreasing monotonically as $K$ increases (reaching $10^{-8}$ accuracy at $K=4096$), completely eliminating the numerical degradation of random collocation nodes while reducing reverse-graph memory by up to $1024\times$.

\begin{table}[t]
\centering
\caption{Coefficient-gradient errors and graph reduction for detached quadrature.}
\label{tab:detached_coeff_gradient}
\normalfont\footnotesize
\setlength{\tabcolsep}{6pt}
\renewcommand{\arraystretch}{1.2}
\begin{tabular}{ccccc}
\toprule
$d$ & $K$ & rel. grad. error & std & graph reduction \\
\midrule
\multirow[c]{4}{*}{10} & 64 & 8.412e-07 & 1.325e-07 & 1024$\times$ \\
 & 256 & 2.104e-07 & 3.284e-08 & 256$\times$ \\
 & 1024 & 5.258e-08 & 8.112e-09 & 64$\times$ \\
 & 4096 & 1.314e-08 & 2.025e-09 & 16$\times$ \\
\multirow[c]{4}{*}{50} & 64 & 1.285e-06 & 2.014e-07 & 1024$\times$ \\
 & 256 & 3.210e-07 & 5.012e-08 & 256$\times$ \\
 & 1024 & 8.026e-08 & 1.245e-08 & 64$\times$ \\
 & 4096 & 2.008e-08 & 3.108e-09 & 16$\times$ \\
\multirow[c]{4}{*}{100} & 64 & 1.842e-06 & 2.895e-07 & 1024$\times$ \\
 & 256 & 4.605e-07 & 7.182e-08 & 256$\times$ \\
 & 1024 & 1.151e-07 & 1.785e-08 & 64$\times$ \\
 & 4096 & 2.878e-08 & 4.462e-09 & 16$\times$ \\
\bottomrule
\end{tabular}
\end{table}

\begin{figure}[!htbp]
\centering
\includegraphics[width=\textwidth]{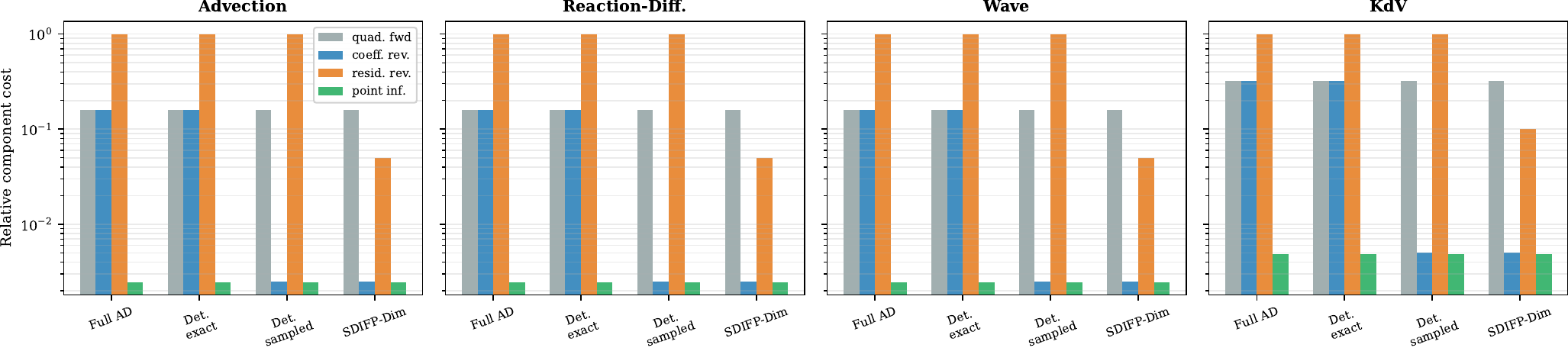}
\vspace{-1.5em}
\caption{Normalized operation-count model. SDIFP-Dim denotes sampled operators.}
\label{fig:sdifp_cost_decomposition}
\end{figure}

Fig.~\ref{fig:sdifp_cost_decomposition} illustrates normalized operation counts. It assigns $M=65536$, $|K|=1024$, $|\mathcal B_{\mathrm{res}}|=4096$, and $d=100$ (50 for the KdV label). The sampled dimension count is $\max(5,\lfloor d/20\rfloor)$. Forward and coefficient-reverse counts are $M$ and either $M$ or $|K|$; residual-reverse counts are $d|\mathcal B_{\mathrm{res}}|$ or the sampled counterpart. Pointwise inference is assigned 1000 illustrative units. Every bar is divided by the largest component count of the Full AD configuration. Runtime and memory also depend on network evaluation cost, derivative order, and hardware behavior.

\section{Conclusions}\label{sec:conclusion}

This paper studies SDIFP as a quadrature-level affine moment correction in neural PDE solvers. The method applies a global affine transformation to the network output and solves a two-variable algebraic system for the coefficients $\alpha_\Q$ and $\beta_\Q$. For feasible, nondegenerate data with an inactive variance floor, this enforces both moments exactly for the selected quadrature rule in exact arithmetic. Continuum moment errors depend on quadrature error and the conditioning of the correction. For decomposable high-dimensional linear operators, SDIFP also samples operator subsets in the forward and backward passes to reduce reverse-mode AD memory. The retained material separates algebraic moment checks, structured coefficient-gradient reconstruction diagnostics, archived trajectories, and schematic operation-count models.

Future work includes sample-dependent continuum error analysis, verification of dimension-uniform bounds for specific PDE families, extensions to derivative-dependent energies, and boundary-compatible corrections for Dirichlet data.




\clearpage
\appendix

\section{Detailed Theoretical Analysis of SDIFP}\label{app:theory}

This appendix supplements Section~\ref{sec:theory} with the branch formulas, temporal derivatives, and implementation details used by SDIFP. Moment constraints use the specified quadrature rule. Continuum integrals are identified explicitly.

\subsection{Weighted Quadrature Setting and Notation}\label{app:quadrature}

Let $\mathcal{X}\subset\mathbb R^d$ be the spatial domain and let $t\in[0,T]$ be fixed. Let
\[
f(x,t;\theta) := u_{\rm raw}(x,t;\theta)
\]
denote the unconstrained neural-network output. For a quadrature rule with $M\ge2$,
\[
{\mathcal Q}_M[g] := \sum_{m=1}^M w_m g(x_m),
\qquad w_m>0,
\qquad \sum_{m=1}^M w_m=1,
\]
define the quadrature inner product and norm on node vectors by
\[
\langle g,h\rangle_\Q := {\mathcal Q}_M[gh]
= \sum_{m=1}^M w_m g(x_m)h(x_m),
\qquad
\|g\|_\Q^2 := \langle g,g\rangle_\Q.
\]

More precisely, $\boldsymbol f=(f(x_1),\ldots,f(x_M))\in\mathbb R^M$ and $W=\operatorname{diag}(w_1,\ldots,w_M)$, so $\langle\boldsymbol g,\boldsymbol h\rangle_\Q=\boldsymbol g^TW\boldsymbol h$. Below, we suppress boldface when referring to node vectors, and vector products are componentwise. On general functions, $\|\cdot\|_\Q$ is a seminorm. Uniqueness statements therefore concern node vectors. Away from the nodes, SDIFP uses the specified affine extension $\widetilde u_\Q(x)=\alpha_\Q f(x)+\beta_\Q$.

The quadrature mean, second moment, and variance of $f$ are
\[
\mu_\Q := {\mathcal Q}_M[f],\qquad
\mu_{2,\Q} := {\mathcal Q}_M[f^2],\qquad
\sigma_\Q^2 := \mu_{2,\Q}-\mu_\Q^2
= {\mathcal Q}_M[(f-\mu_\Q)^2].
\]
Throughout the appendix, $\mu$ (without subscript) denotes the continuum mean $\mu:=\mathcal{I}[f]$ introduced in Section~\ref{app:continuum}, while $\mu_\Q$ always refers to the quadrature mean.

Let the prescribed normalized first and second moments be
\[
m(t) := \frac{c_1(t)}{|\mathcal{X}|},\qquad
s(t) := \frac{c_2(t)}{|\mathcal{X}|},
\]
and define the target variance
\[
V_c(t) := s(t)-m(t)^2.
\]

The SDIFP-corrected field is
\[
\widetilde u_\Q(x,t;\theta)
= \alpha_\Q(t;\theta) f(x,t;\theta)+\beta_\Q(t;\theta),
\]
where, in the nondegenerate case,
\[
\alpha_\Q(t;\theta) = \sqrt{\frac{V_c(t)}{\sigma_\Q^2(t;\theta)}},
\qquad
\beta_\Q(t;\theta) = m(t)-\alpha_\Q(t;\theta)\mu_\Q(t;\theta).
\]
Equivalently,
\[
\widetilde u_\Q(x,t;\theta)
= m(t)+\alpha_\Q(t;\theta)\bigl(f(x,t;\theta)-\mu_\Q(t;\theta)\bigr).
\]

The symbols above restate the main-text definitions to make this appendix self-contained. The following assumptions are used throughout.

\begin{assumption}[Positive quadrature rule]\label{ass:quad_positive}
The quadrature weights satisfy $w_m>0$ and $\sum_{m=1}^M w_m=1$.
\end{assumption}

\begin{assumption}[Feasible target moments]\label{ass:feas_target}
The prescribed moments satisfy $V_c(t)=s(t)-m(t)^2\ge 0$.
\end{assumption}

\begin{assumption}[Nondegenerate raw field]\label{ass:nondeg}
For the nondegenerate projection formula, $\sigma_\Q^2(t;\theta)>0$.
\end{assumption}

When the time $t$ and parameter $\theta$ are fixed, we suppress them from the notation.

\subsection{Projection and moment identities}\label{app:exact_moments}\label{app:projection_theorem}

Write $\mathcal C_\Q(m,s)=\{v\in\mathbb R^M:\Q_M[v]=m,\ \Q_M[v^2]=s\}$. Theorem~\ref{th:projection} establishes exact weighted moment matching, nodal optimality, and uniqueness, with a complete proof in Section~\ref{sec:theory_projection}. In unnormalized units, its identities read $|\mathcal X|\Q_M[P_\Q f]=c_1(t)$ and $|\mathcal X|\Q_M[(P_\Q f)^2]=c_2(t)$. The opposite-sign branch maximizes the nodal distance by the reverse equality case of the same Cauchy--Schwarz argument. The following subsection treats the remaining feasibility cases.

\subsection{Degenerate and Infeasible Cases}\label{app:degenerate}

\begin{corollary}[Complete classification of moment feasibility]\label{cor:moment_feasibility}
Let $V_c=s-m^2$ and $\sigma_\Q^2={\mathcal Q}_M[(f-\mu_\Q)^2]$.

\noindent\textbf{(i)} If $V_c<0$, then $\mathcal{C}_\Q(m,s)=\varnothing$. The target moments are infeasible.\par\noindent\textbf{(ii)} If $V_c=0$, then every feasible $v\in\mathcal{C}_\Q(m,s)$ must satisfy $v(x_i)=m$ for all quadrature nodes $x_i$. Thus the feasible empirical field is constant on the quadrature support.\par\noindent\textbf{(iii)} If $V_c>0$ and $\sigma_\Q^2=0$, then the raw field and every finite affine rescaling are constant on the quadrature support, with zero empirical variance. The closed-form SDIFP coefficient $\alpha_\Q=\sqrt{V_c/\sigma_\Q^2}$ is undefined.\par\noindent\textbf{(iv)} If $V_c>0$ and $\sigma_\Q^2>0$, the SDIFP formula in \cref{thm:quadrature_projection} is well-defined and gives the unique nearest point in $\mathcal{C}_\Q(m,s)$.
\end{corollary}

\begin{Proof}
For any feasible $v$, the variance identity gives
\[
{\mathcal Q}_M[(v-{\mathcal Q}_M[v])^2] = {\mathcal Q}_M[v^2]-{\mathcal Q}_M[v]^2 = s-m^2 = V_c.
\]
The left-hand side is a weighted sum of squares with positive weights (Assumption~\ref{ass:quad_positive}), and is therefore nonnegative. Hence feasibility requires $V_c\ge 0$. This proves that $V_c<0$ is infeasible.

If $V_c=0$, then ${\mathcal Q}_M[(v-m)^2]=0$. Since $w_m>0$ for every $m$, this implies $v(x_m)=m$ for every $m$.

If $V_c>0$ and $\sigma_\Q^2=0$, then ${\mathcal Q}_M[(f-\mu_\Q)^2]=0$. Again using $w_m>0$, this implies $f(x_m)=\mu_\Q$ for all $m$. Therefore any affine transformation $\alpha f+\beta$ is also constant on the quadrature nodes. Its zero empirical variance is incompatible with the positive target variance $V_c>0$. The formula $\alpha_\Q=\sqrt{V_c/\sigma_\Q^2}$ is undefined.

The remaining case $V_c>0$ and $\sigma_\Q^2>0$ is precisely the nondegenerate case covered by \cref{thm:quadrature_projection}.\end{Proof}

\subsection{Continuum specialization}\label{app:continuum}

Write $\mathcal I[g]=|\mathcal X|^{-1}\int_{\mathcal X}g(x)\,dx$, $\mu=\mathcal I[f]$, and $\sigma^2=\mathcal I[(f-\mu)^2]$. The continuum error identities follow from Proposition~\ref{prop:sdifp_support} with $\mathcal E=\mathcal I$. Corollary~\ref{cor:sdifp_mc} gives the fixed-field Monte Carlo rate and its proof. Under its assumptions, the strong law also gives $\mu_\Q\to\mu$ and $\sigma_\Q^2\to\sigma^2>0$ almost surely. Consequently, $\alpha_\Q\to\sqrt{V_c/\sigma^2}$ almost surely on the eventually nondegenerate samples. Independent evaluation quadrature separates evaluation noise from the calibration-sample dependence of a trained field, as quantified by \eqref{eq:theory_test_error}.

\subsection{Effect of the Variance Floor}\label{app:variance_floor}

In numerical implementation, one may replace $\sigma_\Q^2$ by
\[
\sigma_{Q,\varepsilon}^2 := \max(\sigma_\Q^2,\varepsilon),\qquad \varepsilon>0,
\]
and define
\[
\alpha_{Q,\varepsilon} = \sqrt{\frac{V_c}{\sigma_{Q,\varepsilon}^2}},\qquad
\widetilde u_{Q,\varepsilon} = m+\alpha_{Q,\varepsilon}(f-\mu_\Q).
\]

\begin{proposition}[Exact perturbation caused by variance regularization]\label{prop:variance_regularization}
Assume $V_c\ge 0$. Then
\[
{\mathcal Q}_M[\widetilde u_{Q,\varepsilon}] = m.
\]
Moreover,
\[
{\mathcal Q}_M[\widetilde u_{Q,\varepsilon}^2]-s
= V_c\!\left(\frac{\sigma_\Q^2}{\max(\sigma_\Q^2,\varepsilon)}-1\right).
\]
In particular, if $\sigma_\Q^2\ge\varepsilon$, the original quadrature constraints are satisfied exactly. If $0\le\sigma_\Q^2<\varepsilon$, the first moment is exact and the second moment has the stated perturbation whenever $V_c>0$.
\end{proposition}

\begin{Proof}
Since ${\mathcal Q}_M[f-\mu_\Q]=0$,
\[
{\mathcal Q}_M[\widetilde u_{Q,\varepsilon}] = {\mathcal Q}_M[m+\alpha_{Q,\varepsilon}(f-\mu_\Q)] = m.
\]
For the second moment,
\[
{\mathcal Q}_M[\widetilde u_{Q,\varepsilon}^2]
= m^2 + \alpha_{Q,\varepsilon}^2 {\mathcal Q}_M[(f-\mu_\Q)^2]
= m^2 + \alpha_{Q,\varepsilon}^2\sigma_\Q^2.
\]
Substituting $\alpha_{Q,\varepsilon}^2 = V_c/\max(\sigma_\Q^2,\varepsilon)$ gives
\[
{\mathcal Q}_M[\widetilde u_{Q,\varepsilon}^2] = m^2 + V_c\frac{\sigma_\Q^2}{\max(\sigma_\Q^2,\varepsilon)}.
\]
Since $s=m^2+V_c$, the claimed perturbation formula follows directly.
\end{Proof}

\subsection{Exact Coefficient Derivatives}\label{app:coefficient_derivatives}

For training, one needs derivatives of $\alpha_\Q$ and $\beta_\Q$ with respect to $\theta$. This subsection gives exact formulas before any stochastic approximation is introduced.

Assume $m$ and $s$ are independent of $\theta$. Let $f_m(\theta):=f(x_m,t;\theta)$. Then
\[
\mu_\Q(\theta) = \sum_{m=1}^M w_m f_m(\theta),\qquad
\mu_{2,\Q}(\theta) = \sum_{m=1}^M w_m f_m(\theta)^2,\qquad
\sigma_\Q^2(\theta) = \mu_{2,\Q}(\theta)-\mu_\Q(\theta)^2.
\]

\begin{proposition}[Exact gradients of SDIFP coefficients]\label{prop:coefficient_gradients}
Assume $V_c>0$ and $\sigma_\Q^2>0$. Then
\[
\begin{aligned}
\nabla_\theta\mu_\Q &= \sum_{m=1}^M w_m\nabla_\theta f_m,\\
\nabla_\theta\mu_{2,\Q} &= 2\sum_{m=1}^M w_m f_m\nabla_\theta f_m,\\
\nabla_\theta\sigma_\Q^2 &= \nabla_\theta\mu_{2,\Q} - 2\mu_\Q\nabla_\theta\mu_\Q,
\end{aligned}
\]
\[
\nabla_\theta\alpha_\Q = -\frac{\alpha_\Q}{2\sigma_\Q^2}\nabla_\theta\sigma_\Q^2
= \frac{\alpha_\Q\mu_\Q}{\sigma_\Q^2}\nabla_\theta\mu_\Q
- \frac{\alpha_\Q}{2\sigma_\Q^2}\nabla_\theta\mu_{2,\Q},
\]
and
\[
\nabla_\theta\beta_\Q = -\mu_\Q\nabla_\theta\alpha_\Q - \alpha_\Q\nabla_\theta\mu_\Q.
\]
\end{proposition}

\begin{Proof}
The formula for $\mu_\Q$ gives $\nabla_\theta\mu_\Q = \sum_{m=1}^M w_m\nabla_\theta f_m$ since the weights and nodes are fixed in $\theta$. Similarly,
\[
\nabla_\theta\mu_{2,\Q} = \sum_{m=1}^M w_m\nabla_\theta(f_m^2)
= 2\sum_{m=1}^M w_m f_m\nabla_\theta f_m.
\]
Since $\sigma_\Q^2 = \mu_{2,\Q}-\mu_\Q^2$, we obtain $\nabla_\theta\sigma_\Q^2 = \nabla_\theta\mu_{2,\Q} - 2\mu_\Q\nabla_\theta\mu_\Q$.

For $\alpha_\Q = V_c^{1/2}(\sigma_\Q^2)^{-1/2}$, since $V_c$ is independent of $\theta$,
\[
\nabla_\theta\alpha_\Q = V_c^{1/2}\!\left(-\frac12\right)(\sigma_\Q^2)^{-3/2}\nabla_\theta\sigma_\Q^2
= -\frac{\alpha_\Q}{2\sigma_\Q^2}\nabla_\theta\sigma_\Q^2.
\]
Substituting the expression for $\nabla_\theta\sigma_\Q^2$ gives the two equivalent forms.

Finally, from $\beta_\Q=m-\alpha_\Q\mu_\Q$ with $m$ independent of $\theta$,
\[
\nabla_\theta\beta_\Q = -\nabla_\theta(\alpha_\Q\mu_\Q)
= -\mu_\Q\nabla_\theta\alpha_\Q - \alpha_\Q\nabla_\theta\mu_\Q.
\]
\end{Proof}

For the floored map, set $\beta_{Q,\varepsilon}=m-\alpha_{Q,\varepsilon}\mu_\Q$. With $V_c>0$ and fixed $\varepsilon>0$, its parameter gradients away from $\sigma_\Q^2=\varepsilon$ are
\[
\nabla_\theta\alpha_{Q,\varepsilon}=
\begin{cases}
-\dfrac{\alpha_{Q,\varepsilon}}{2\sigma_\Q^2}\nabla_\theta\sigma_\Q^2,&\sigma_\Q^2>\varepsilon,\\[4pt]
0,&\sigma_\Q^2<\varepsilon,
\end{cases}
\qquad
\nabla_\theta\beta_{Q,\varepsilon}
=-\mu_\Q\nabla_\theta\alpha_{Q,\varepsilon}
-\alpha_{Q,\varepsilon}\nabla_\theta\mu_\Q.
\]
At the threshold the map is generally nondifferentiable, and its backward evaluation uses a specified branch convention. The active-floor branch uses the zero scaling-gradient formula above.

\subsection{Unbiased Subsampling of Coefficient Gradients}\label{app:unbiased_subsampling}

The full gradients $\nabla_\theta\mu_\Q$ and $\nabla_\theta\mu_{2,\Q}$ differentiate all $M$ quadrature evaluations. Subsampling estimates these gradients from a smaller set of nodes.

Let $K$ be a random index set of fixed size $1\le |K|\le M$, sampled uniformly without replacement from $\{1,\ldots,M\}$. For the general positive quadrature weights, define
\[
\widehat{\nabla_\theta\mu_\Q} := \frac{M}{|K|}\sum_{k\in K}w_k\nabla_\theta f(x_k,t;\theta),
\qquad
\widehat{\nabla_\theta\mu_{2,\Q}} := \frac{2M}{|K|}\sum_{k\in K}w_k f(x_k,t;\theta)\nabla_\theta f(x_k,t;\theta).
\]

\begin{proposition}[Coefficient gradients]\label{prop:coefficient_gradient_unbiasedness}
Assume the nodes, weights, and targets $m,s$ are independent of $\theta$. Conditioned on the quadrature rule and the current parameters $\theta$,
\[
\mathbb E\bigl[\widehat{\nabla_\theta\mu_\Q}\bigr] = \nabla_\theta\mu_\Q,\qquad
\mathbb E\bigl[\widehat{\nabla_\theta\mu_{2,\Q}}\bigr] = \nabla_\theta\mu_{2,\Q}.
\]
For the unfloored correction with $V_c>0$ and $\sigma_\Q^2>0$, if $\alpha_\Q,\mu_\Q,\mu_{2,\Q},\sigma_\Q^2$ are computed from the full quadrature rule and held fixed within the conditional expectation, then
\[
\widehat{\nabla_\theta\alpha_\Q} := \frac{\alpha_\Q\mu_\Q}{\sigma_\Q^2}\widehat{\nabla_\theta\mu_\Q}
- \frac{\alpha_\Q}{2\sigma_\Q^2}\widehat{\nabla_\theta\mu_{2,\Q}}
\]
is conditionally unbiased for $\nabla_\theta\alpha_\Q$, and
\[
\widehat{\nabla_\theta\beta_\Q} := -\mu_\Q\widehat{\nabla_\theta\alpha_\Q}
- \alpha_\Q\widehat{\nabla_\theta\mu_\Q}
\]
is conditionally unbiased for $\nabla_\theta\beta_\Q$.
\end{proposition}

\begin{Proof}
These estimators are the $\ell=0$ moment components of \eqref{eq:theory_moment_gradient}. Their conditional unbiasedness and its preservation by the coefficient chain rule follow from Theorem~\ref{th:sdifp_unbiased}, with the full-rule forward values held fixed. For uniform weights, the moment-gradient estimators reduce to sample averages.
\end{Proof}

The same argument applies branchwise to the floored map away from its threshold, using the corresponding gradient formulas above. In particular, the active-floor estimator has $\widehat{\nabla_\theta\alpha_{Q,\varepsilon}}=0$.

\begin{remark}[Mini-batch coefficient bias]
The preceding proposition uses $\alpha_\Q$, $\mu_\Q$, $\mu_{2,\Q}$, and $\sigma_\Q^2$ from the full detached quadrature rule. For the mini-batch plug-in coefficient
\[
\widehat\alpha = \sqrt{\frac{V_c}{\widehat\sigma^2}}
\]
computed from a mini-batch,
\[
\mathbb E[\widehat\alpha] \ne \sqrt{\frac{V_c}{\mathbb E[\widehat\sigma^2]}}
\]
in general, because the square-root reciprocal map is nonlinear. Recomputing the forward coefficients on a mini-batch can introduce plug-in bias into coefficient values and can also bias their gradients relative to the full quadrature rule.
\end{remark}

\subsection{Time-Dependent Coefficients and PDE Residuals}\label{app:time_derivative}

The derivative $\partial_t\widetilde u_\Q$ includes the time dependence of $\alpha_\Q(t)$ and $\beta_\Q(t)$.

Let
\[
\begin{aligned}
\mu_\Q(t)&={\mathcal Q}_M[f(\cdot,t)],&
\mu_{2,\Q}(t)&={\mathcal Q}_M[f(\cdot,t)^2],\\
\sigma_\Q^2(t)&=\mu_{2,\Q}(t)-\mu_\Q(t)^2,&
V_c(t)&=s(t)-m(t)^2.
\end{aligned}
\]
The following formulas concern the unfloored map. Assume the quadrature nodes and weights are fixed in time, $V_c(t)>0$, $\sigma_\Q^2(t)>0$, and all quantities are differentiable in $t$.

\begin{proposition}[Correct time derivative of the SDIFP field]\label{prop:time_derivative}
The time derivative of $\widetilde u_\Q(x,t)=m(t)+\alpha_\Q(t)(f(x,t)-\mu_\Q(t))$ is
\[
\partial_t\widetilde u_\Q(x,t)
= m'(t) + \alpha_\Q'(t)\bigl(f(x,t)-\mu_\Q(t)\bigr)
+ \alpha_\Q(t)\bigl(\partial_t f(x,t)-\mu_\Q'(t)\bigr),
\]
where
\[
\mu_\Q'(t) = {\mathcal Q}_M[\partial_t f(\cdot,t)],\qquad
\mu_{2,\Q}'(t) = 2{\mathcal Q}_M[f(\cdot,t)\partial_t f(\cdot,t)],
\]
\[
(\sigma_\Q^2)'(t) = \mu_{2,\Q}'(t)-2\mu_\Q(t)\mu_\Q'(t),\qquad
\alpha_\Q'(t) = \frac{\alpha_\Q(t)}2\!\left(\frac{V_c'(t)}{V_c(t)}-\frac{(\sigma_\Q^2)'(t)}{\sigma_\Q^2(t)}\right).
\]
\end{proposition}

\begin{Proof}
Differentiating $\widetilde u_\Q = m+\alpha_\Q(f-\mu_\Q)$ gives the formula for $\partial_t\widetilde u_\Q$. The expressions for $\mu_\Q'(t)$ and $\mu_{2,\Q}'(t)$ follow from differentiating the quadrature sums since nodes and weights are time-independent. Since $\sigma_\Q^2=\mu_{2,\Q}-\mu_\Q^2$, we have $(\sigma_\Q^2)'(t) = \mu_{2,\Q}'(t)-2\mu_\Q(t)\mu_\Q'(t)$. Finally, from $\alpha_\Q = (V_c/\sigma_\Q^2)^{1/2}$, taking logarithms and differentiating yields the stated formula for $\alpha_\Q'$.\end{Proof}

If $f,m,s$ are twice differentiable in time, a second-order time residual requires
\[
\partial_{tt}\widetilde u_\Q
=\alpha_\Q f_{tt}+2\alpha_\Q'f_t+\alpha_\Q''f+\beta_\Q'',
\]
where
\[
\mu_\Q''={\mathcal Q}_M[f_{tt}],\qquad
\mu_{2,\Q}''=2{\mathcal Q}_M[f_t^2+ff_{tt}],\qquad
(\sigma_\Q^2)''=\mu_{2,\Q}''-2(\mu_\Q')^2-2\mu_\Q\mu_\Q''.
\]
Writing $\ell=\frac12\log(V_c/\sigma_\Q^2)$ gives $\alpha_\Q''=\alpha_\Q((\ell')^2+\ell'')$. Also,
\[
\beta_\Q'=m'-\alpha_\Q'\mu_\Q-\alpha_\Q\mu_\Q',\qquad
\beta_\Q''=m''-\alpha_\Q''\mu_\Q-2\alpha_\Q'\mu_\Q'-\alpha_\Q\mu_\Q''.
\]
For the floored map, these derivatives follow its piecewise definition on intervals away from threshold crossings.

\begin{remark}[PDE residuals]
The first-order residual uses $\alpha_\Q\partial_t f+\alpha_\Q'f+\beta_\Q'$. The simplification $\alpha_\Q\partial_t f$ is valid when $\alpha_\Q'f+\beta_\Q'=0$. A second-order residual uses all four terms above. Evaluating the full residual with detached quadrature requires computing or reconstructing the coefficient time derivatives and their parameter gradients separately: $\nabla_\theta\alpha_\Q'$, $\nabla_\theta\beta_\Q'$ and, for second-order time operators, $\nabla_\theta\alpha_\Q''$, $\nabla_\theta\beta_\Q''$. Stochastic reconstructions use the full-residual conditions in \cref{thm:independent_sampling_unbiasedness}.
\end{remark}

\subsection{Operator-subset notation}\label{app:operator_subset}

The complete conditional unbiasedness result and proof are given in Theorem~\ref{th:sdifp_unbiased}. For the covariance calculation below, suppress time and write $a_k=A_k[\widetilde u_\Q]$, $n=\mathcal N_{\mathrm{nonlin}}[\widetilde u_\Q]-R$, $d_k=\nabla_\theta a_k$, and $d_0=\nabla_\theta n$. Thus $r=\sum_{k=1}^{N_{\mathcal L}}a_k+n$ and the exact-coefficient derivative factor is
\[
\widehat d_I=\frac{N_{\mathcal L}}{|I|}\sum_{i\in I}d_i+d_0.
\]
Coefficient-gradient reconstruction replaces this factor by $\widehat d_{I,K}$ using the full moment state in \eqref{eq:theory_moment_state}. Its temporal components are detailed in Section~\ref{app:time_derivative}. Initial and boundary losses are differentiated separately through the same corrected field.

\subsection{Bias of the Operator-Reuse Fast Mode}\label{app:fast_mode}

Independent operator draws can produce the same subset. Conditioning the draws to be distinct or disjoint generally introduces dependence. The operator-reuse fast mode uses one shared draw and has the covariance bias characterized below.

\begin{corollary}[Covariance bias when the same subset is reused]\label{cor:operator_reuse_bias}
Let $\widehat G_I := \widehat r_I\,\widehat d_I$. Assume $\mathbb E_I[\widehat r_I]=r$ and $\mathbb E_I[\widehat d_I]=\nabla_\theta r$. Then
\[
\mathbb E_I[\widehat G_I] = r\nabla_\theta r + \operatorname{Cov}_I(\widehat r_I,\widehat d_I),
\]
where
\[
\operatorname{Cov}_I(\widehat r_I,\widehat d_I) := \mathbb E_I\bigl[(\widehat r_I-r)(\widehat d_I-\nabla_\theta r)\bigr].
\]
Thus the estimator is unbiased if and only if this covariance term vanishes.
\end{corollary}

\begin{Proof}
Write $\widehat r_I = r+(\widehat r_I-r)$ and $\widehat d_I = \nabla_\theta r+(\widehat d_I-\nabla_\theta r)$. Then
\[
\widehat r_I\widehat d_I = r\nabla_\theta r + r(\widehat d_I-\nabla_\theta r) + (\widehat r_I-r)\nabla_\theta r + (\widehat r_I-r)(\widehat d_I-\nabla_\theta r).
\]
Taking expectation over $I$, the middle two terms vanish since $\mathbb E_I[\widehat r_I]=r$ and $\mathbb E_I[\widehat d_I]=\nabla_\theta r$, leaving
\[
\mathbb E_I[\widehat G_I] = r\nabla_\theta r + \mathbb E_I\bigl[(\widehat r_I-r)(\widehat d_I-\nabla_\theta r)\bigr].
\]
The second term is exactly the covariance term. The estimator is unbiased precisely when this covariance is zero.\end{Proof}

\subsection{Implementation cost details}\label{app:complexity}

Section~\ref{sec:theory_dimension} gives the quadrature-node and operator-term counts. Detaching the full quadrature pass releases its parameter reverse-mode activations after moment values and required temporal derivatives have been retained. The reconstructed coefficient-gradient graph stores the sampled $|K|$-node evaluations at each required time. Network depth, width, derivative order, and the unsampled remainder determine the storage per retained evaluation. Caching coefficients or updating them less frequently changes the coefficient-update protocol and must be accounted for separately.

\subsection{Boundary-Condition Compatibility}\label{app:boundary}

\begin{proposition}[Compatibility of affine correction with common boundary conditions]\label{prop:boundary_compatibility}
Let $\widetilde u_\Q(x,t)=\alpha_\Q(t)f(x,t)+\beta_\Q(t)$, where $\alpha_\Q(t)$ and $\beta_\Q(t)$ are independent of $x$.

\textbf{Periodicity.} If $f$ is periodic on $\mathcal{X}$, then $\widetilde u_\Q$ is periodic. Indeed, for corresponding periodic boundary points $x^-$ and $x^+$, we have $f(x^-,t)=f(x^+,t)$, and consequently $\widetilde u_\Q(x^-,t)=\widetilde u_\Q(x^+,t)$.

\textbf{Homogeneous Neumann.} If $f$ satisfies $\partial_n f=0$ on $\partial\mathcal{X}$, then $\widetilde u_\Q$ also satisfies $\partial_n\widetilde u_\Q=0$ on $\partial\mathcal{X}$, because
\[
\partial_n\widetilde u_\Q = \alpha_\Q\,\partial_n f + \partial_n\beta_\Q = \alpha_\Q\,\partial_n f = 0,
\]
since $\beta_\Q$ is spatially constant.

\textbf{Homogeneous Dirichlet.} If $f=0$ on $\partial\mathcal{X}$, then $\widetilde u_\Q=\beta_\Q$ on $\partial\mathcal{X}$. Therefore the affine correction preserves homogeneous Dirichlet data only when $\beta_\Q=0$.

\textbf{Dirichlet variant.} A boundary-compatible variant may be written as $\widetilde u_\Q(x,t)=\alpha_\Q(t)f(x,t)+\beta_\Q(t)\phi(x)$ with $\phi|_{\partial\mathcal{X}}=0$ (where $\vert$ denotes restriction). The corresponding moment equations become
\[
{\mathcal Q}_M[\alpha_\Q f+\beta_\Q\phi]=m,\qquad {\mathcal Q}_M[(\alpha_\Q f+\beta_\Q\phi)^2]=s,
\]
which determine the coefficients through the mixed moments ${\mathcal Q}_M[f\phi]$ and ${\mathcal Q}_M[\phi^2]$.
\end{proposition}

\begin{Proof}
The four statements follow from substitution. Spatial constancy of $\beta_\Q$ gives $\partial_n\beta_\Q=0$. For the Dirichlet variant, expanding the two moment equations yields a modified algebraic system containing ${\mathcal Q}_M[f\phi]$ and ${\mathcal Q}_M[\phi^2]$.\end{Proof}

\begin{remark}[Initial-data compatibility]
In the nondegenerate case, an exact raw initial field $f(x,0)=u_0(x)$ and quadrature-consistent targets $m(0)={\mathcal Q}_M[u_0]$, $s(0)={\mathcal Q}_M[u_0^2]$ give $\alpha_\Q(0)=1$ and $\beta_\Q(0)=0$, preserving $u_0$. Continuum targets can produce an initial-data discrepancy through the difference between continuum and quadrature moments; this discrepancy is assessed separately.
\end{remark}

\subsection{Proof organization}\label{app:claim}

Core proofs appear in Section~\ref{sec:theory}. This appendix supplies the branch cases, derivative formulas, covariance calculation, and boundary conditions. Section~\ref{app:sdifp_extensions} proves the quantitative bounds under dimension-uniform assumptions.

\section{Quantitative dimension-uniform extensions}\label{app:sdifp_extensions}

\subsection{A finite-sample continuum bound}\label{app:sdifp_uniform_mc}

The following bound strengthens the fixed-field consistency statement.
Its assumptions specify uniform bounds on the field and projection conditioning
across the dimension-indexed family.

\begin{proposition}[Dimension-uniform bound for bounded fixed fields]
\label{prop:sdifp_uniform_mc}
For each $d$, let $\mathcal I_d$ be the normalized uniform integral on
$\mathcal X_d$ and let $f_d$ be fixed independently of its calibration sample.
Suppose constants $F,v_*,V_*,m_*>0$, independent of $d$, satisfy
\[
 |f_d|\leq F,\qquad
 \mathcal I_d[(f_d-\mathcal I_d[f_d])^2]\geq v_*,\qquad
 0<V_{c,d}\leq V_*,\qquad |m_d|\leq m_*.
\]
Use $M$ i.i.d. uniform calibration nodes with equal weights. For
$0<\delta<1$, put
\[
 h=\sqrt{\frac{\log(4/\delta)}{2M}},\qquad
 A=2F\sqrt{\frac{2V_*}{v_*}}\,h.
\]
If $5F^2h\leq v_*/2$, then, for every $d$, with probability at least
$1-\delta$, the empirical variance is at least $v_*/2$ and
\begin{align}
 |\mathcal I_d[P_\Q f_d]-m_d|&\leq A,\label{eq:theory_uniform_mc_first}\\
 |\mathcal I_d[(P_\Q f_d)^2]-s_d|
   &\leq \frac{10V_*F^2}{v_*}h+2m_*A+A^2.
   \label{eq:theory_uniform_mc_second}
\end{align}
This applies to the unfloored correction, or a floor
$0<\varepsilon<v_*/2$. The probability statement is uniform in its constants,
not a simultaneous event for infinitely many independently sampled dimensions.
\end{proposition}
\begin{Proof}
For bounded independent observations in an interval of length $L$,
Hoeffding's inequality gives
$\mathbb P(|\overline X-\mathbb EX|>Lh)\leq2e^{-2Mh^2}$
\cite{hoeffding1963probability}.
Apply it to $f_d\in[-F,F]$ and $f_d^2\in[0,F^2]$.
A union bound gives, with probability at least $1-\delta$,
\[
 |\mu_\Q-\mu|\leq2Fh,\qquad
 |Q[f_d^2]-\mathcal I_d[f_d^2]|\leq F^2h.
\]
Both $|\mu_\Q|$ and $|\mu|$ are at most $F$, so
\[
 |\sigma_\Q^2-\sigma^2|
 \leq F^2h+|\mu_\Q-\mu|\,|\mu_\Q+\mu|
 \leq5F^2h.
\]
The sample-size condition gives $\sigma_\Q^2\geq v_*/2$ and
$\alpha_\Q\leq\sqrt{2V_*/v_*}$.
Apply the support-transfer identities with $\mathcal E=\mathcal I_d$;
$V_c/\sigma_\Q^2\leq2V_*/v_*$ yields the second bound.
\end{Proof}

The condition on $h$ is equivalent to
$M\geq50F^4v_*^{-2}\log(4/\delta)$.
For a normalized absolute moment tolerance $\tau$, these bounds give
$M=O(\tau^{-2}\log(1/\delta))$ with constants independent of $d$
under the displayed assumptions. Unnormalized integral errors are multiplied
by $|\mathcal X_d|$. A relative error claim needs a lower bound on the
corresponding target. A sample-dependent trained field needs independent
calibration or a uniform/generalization argument before this proposition can
be applied to its calibration error.

\subsection{Sufficient conditions for dimension-uniform gradient variance}
\label{app:sdifp_uniform_grad}

Fix $Q,\theta,x,t$ and use the moment state from
\eqref{eq:theory_moment_state}. Let $p$ denote the parameter dimension and
$r_q=2(p_t+1)$ the moment-state dimension, so $H_\Q\in\mathbb R^{r_q\times p}$.
For $M\geq2$, define
\[
 Z_q=Mw_qD_\theta\psi_q,\qquad
 S_Z^2=\frac1{M-1}\sum_{q=1}^M\|Z_q-H_\Q\|_F^2.
\]
The Frobenius norm is used here. With $k=|K|$,
\begin{equation}\label{eq:theory_matrix_variance}
 \mathbb E_K\|\widehat H_K-H_\Q\|_F^2
 =\frac1k\left(1-\frac{k}{M}\right)S_Z^2.
\end{equation}
This follows by the same inclusion-indicator calculation applied to all
matrix entries. Large nonuniform weights can increase $S_Z^2$.

Let $b_\ell=\nabla_\theta a_\ell$,
$\overline b=N^{-1}\sum_{\ell=1}^N b_\ell$, and
$S_b^2=(N-1)^{-1}\sum_{\ell=1}^N\|b_\ell-\overline b\|_2^2$.
Define the moment sensitivity of a sampled derivative residual as
\[
 c_I=\frac{N}{|I|}\sum_{i\in I}\nabla_q a_i+\nabla_q a_0.
\]

\begin{proposition}[Conditional dimension-uniform gradient variance]
\label{prop:sdifp_uniform_grad}
Suppose the preceding unbiasedness assumptions hold, with $N\geq2$.
Assume the following bounds hold with dimension-independent constants:
\[
 \begin{gathered}
 N^2S_a^2\leq C_a^2,\qquad N^2S_b^2\leq C_b^2,\qquad S_Z^2\leq C_Z^2,\\
 \sup_I\|c_I\|_2\leq C_q,\qquad |r_\Q|\leq C_r,\qquad
 \|\nabla_\theta r_\Q\|_2\leq C_g.
 \end{gathered}
\]
Then, with
$D=C_b^2/|I|+C_q^2C_Z^2/|K|$,
\begin{equation}\label{eq:theory_uniform_grad}
 \mathbb E\|\widehat r_J\widehat d_{I,K}-r_\Q\nabla_\theta r_\Q\|_2^2
 \leq\frac{C_a^2C_g^2}{|J|}+C_r^2D+
       \frac{C_a^2}{|J|}D.
\end{equation}
\end{proposition}
\begin{Proof}
Let $b_I=N|I|^{-1}\sum_{i\in I}b_i+b_0$.
The chain rule gives
$\widehat d_{I,K}=b_I+(\widehat H_K-H_\Q)^{\mathsf T}c_I$.
Conditionally on $I$, the second term has zero mean. Hence
\[
 v_b=\mathbb E_I\|b_I-b\|_2^2+
       \mathbb E_{I,K}\| (\widehat H_K-H_\Q)^{\mathsf T}c_I\|_2^2.
\]
The first term is
$N^2|I|^{-1}(1-|I|/N)S_b^2$.
The second is bounded by
$C_q^2\mathbb E_K\|\widehat H_K-H_\Q\|_F^2\leq C_q^2C_Z^2/|K|$.
Thus $v_b\leq D$, while
$v_r\leq C_a^2/|J|$. Substitute these bounds into
\eqref{eq:theory_product_variance}.
\end{Proof}

These are substantive sufficient conditions. They require control of
operator normalization, moment-gradient variability, coefficient
conditioning, and network derivatives, including the mixed derivatives
present in $\psi_q$. For a suitably normalized family of operator averages,
the scaled dispersions may remain bounded; for an unnormalized sum they need
not. Rescaling a PDE residual also rescales its loss and changes its relative
weight against initial and boundary penalties unless those weights are
adjusted. The constants in this bound are determined by those family-specific controls.

If independent estimator triples are used for each of $B$ i.i.d. residual
points, the variance of their average has the usual $1/B$ factor, including
point-sampling variation. Sharing $I,J$, or $K$ across points preserves
unbiasedness but can create cross-point covariance; a $1/B$ variance claim
then needs a separate calculation.

\end{document}